\documentclass{article} 
\usepackage{collas2023_conference,times}
\usepackage{easyReview}

\usepackage{amsmath,amsfonts,bm}

\def\eqref#1{equation~\ref{#1}}

\def\1{\bm{1}}

\DeclareMathAlphabet{\mathsfit}{\encodingdefault}{\sfdefault}{m}{sl}
\SetMathAlphabet{\mathsfit}{bold}{\encodingdefault}{\sfdefault}{bx}{n}

\DeclareMathOperator*{\argmin}{arg\,min}

\usepackage{hyperref}
\hypersetup{
    colorlinks=true,
    linkcolor=red,
    filecolor=magenta,
    urlcolor=blue,
    citecolor=purple,
    pdftitle={Stabilizing Unsupervised Environment Design with a Learned Adversary},
    pdfpagemode=FullScreen,
    }

\usepackage{algorithm, algorithmic}
\usepackage{float, graphicx}
\usepackage{subfigure}
\usepackage{booktabs}
\usepackage{wrapfig}
\usepackage{rotating}
\usepackage{markdown}
\usepackage{natbib}
\usepackage{easyReview}

\title{Stabilizing Unsupervised Environment Design \\ with a Learned Adversary}

\author{Ishita Mediratta \\
Meta AI \\
\texttt{ishitamed@meta.com} \\
\And
Minqi Jiang\thanks{: Equal Contribution} \\
UCL, Meta AI \\
\And
Jack Parker-Holder$^*$ \\
University of Oxford \\
\AND
Michael Dennis \\
University of California, Berkeley \\
\And
Eugene Vinitsky \\
New York University \\
\And
Tim Rocktäschel \\
UCL
}

\collasfinalcopy 

\begin{document}

\maketitle

\begin{abstract}
A key challenge in training generally-capable agents is the design of training tasks that facilitate broad generalization and robustness to environment variations. This challenge motivates the problem setting of \emph{Unsupervised Environment Design} (UED), whereby a student agent trains on an adaptive distribution of tasks proposed by a teacher agent. A pioneering approach for UED is PAIRED, which uses reinforcement learning (RL) to train a teacher policy to design tasks from scratch, making it possible to directly generate tasks that are adapted to the agent's current capabilities. Despite its strong theoretical backing, PAIRED suffers from a variety of challenges that hinder its practical performance. Thus, state-of-the-art methods currently rely on \emph{curation} and \emph{mutation} rather than \emph{generation} of new tasks. In this work, we investigate several key shortcomings of PAIRED and propose solutions for each shortcoming. As a result, we make it possible for PAIRED to match or exceed state-of-the-art methods, producing robust agents in several established challenging procedurally-generated environments, including a partially-observed maze navigation task and a continuous-control car racing environment. We believe this work motivates a renewed emphasis on UED methods based on learned models that directly generate challenging environments, potentially unlocking more open-ended RL training and, as a result, more general agents.
\end{abstract}

\section{Introduction}
\label{introduction}

Deep reinforcement learning \citep[RL;][]{Sutton1998} has been successfully applied to many challenging domains in recent years ranging from games \citep{alphago, alphazero, alphastar, dota, hu2021off} to real world problems such as controlling nuclear fusion plasma \citep{degrave2022magnetic}. 
Many of these achievements are attributed to techniques like domain randomization and self-play, which provide an adaptive curriculum for training the agents. While these methods have led to impressive results, most successes are limited to single domains, necessitating re-training for each new setting. Instead, there has recently been interest in training more \emph{generally capable} agents~\citep{xland,ada}, which remains a considerable challenge \citep{zhang2018generalizationcont, observational_overfitting}. 

As focus shifts from mastery to generality, emphasis shifts away from designing new agents to focus more on generating sufficiently rich environments. However, manually designing these environments is a tremendous engineering challenge, and even when possible, it is often the case that many instances of the environment are incompatible with training a robust generalist agent. Instead, it may be desirable to automatically discover useful training environments. In light of this, the  \emph{Unsupervised Environment Design} \citep[UED;][]{PAIRED} paradigm has emerged, whereby a \emph{student} agent trains on an adaptive distribution of tasks proposed by a \emph{teacher} which ultimately aids in automatically discovering a curriculum of environments at the frontier of the agent capabilities. 

A pioneering approach for UED is PAIRED, where the teacher is an RL agent trained to maximize \emph{relative regret}, defined as the difference in performance between two student policies, called the protagonist (the student policy) and an antagonist (cooperating with the teacher agent). Using the teacher objective of maximizing regret between the antagonist and protagonist, \citet{PAIRED} showed that the teacher proposes increasingly complex environments, which leads to a robust generalist student.

However, PAIRED suffers from a variety of challenges which hinder performance. The predominant issue is that optimizing the teacher with RL is incredibly challenging and suffers from the following problems:
\begin{itemize}
    \item \textbf{\textit{Non-Stationarity:}} Both student policies, the protagonist and the antagonist, are updating, making the problem \textit{nonstationary}.
    \item \textbf{\textit{Long-term Credit Assignment:}} There is a challenging \textit{credit assignment problem} since the teacher must fully specify an environment before receiving a sparse reward in the form of feedback from the students.
    \item \textbf{\textit{High-dimensionality:}} The \textit{high-dimensionality of the space} can make the task difficult for the RL algorithms used by the teacher and students, leading to sub-optimal performance.
\end{itemize}

Due to the above-mentioned problems, PAIRED empirically tends to perform worse than one would expect it to based on the theoretical guarantees. In this paper, we will show and also try to alleviate the following two problems:

\begin{enumerate}
    \item \textbf{\textit{P1 - Entropy Collapse:}} The adversary (teacher) faces a challenging RL task due to the high-dimensionality of the environment design space, leading to difficulties in exploration and policy entropy collapse for the agents.
    \item \textbf{\textit{P2 - Fall Behind:}} In the PAIRED curriculum, protagonists can still face difficult exploration tasks, leading to sub-optimal performance. We demonstrate one such case in the Car Racing environment where the teacher produces simple levels favoring the antagonist, causing the protagonist to fall behind.
\end{enumerate}

Thus, alternative UED algorithms focus on sampling tasks uniformly at random from the full distribution, relying on \emph{curation} and \emph{mutation} to provide an effective curriculum (see Section \ref{sec:ued_curation}), as opposed to \emph{generation} (guided by the regret metric, which aids in producing meaningful levels based on the agent's current capabilities) in PAIRED. 

In this paper, we seek to investigate the major shortcomings in PAIRED, seeing if PAIRED can be made competitive with state-of-the-art methods. We propose a variety of techniques for stabilizing training by overcoming problems \textbf{\textit{P1, P2}}. In particular, we explore regularization approaches to aid the teacher in better exploration of the level-design space, alternative optimizers and finally behavior distillation across students. 
In each case we provide rigorous empirical analysis and demonstrate the specific mechanisms by which these approaches improve performance. As a result of these variations, we make it possible for PAIRED to match or exceed state-of-the-art methods, producing general agents in challenging procedurally generated environments. We believe this work paves the way for a renewed emphasis on methods that learn to automatically \emph{generate} environments, potentially unlocking more open-ended RL training and as a result more robust, general agents. We have released the accompanying code as part of the existing DCD repository at \href{https://github.com/facebookresearch/dcd}{https://github.com/facebookresearch/dcd}.


\begin{figure}[t!]
\vspace{-5mm}
\centering\subfigure[From UPOMDP to POMDP]{\includegraphics[width=.38\linewidth]{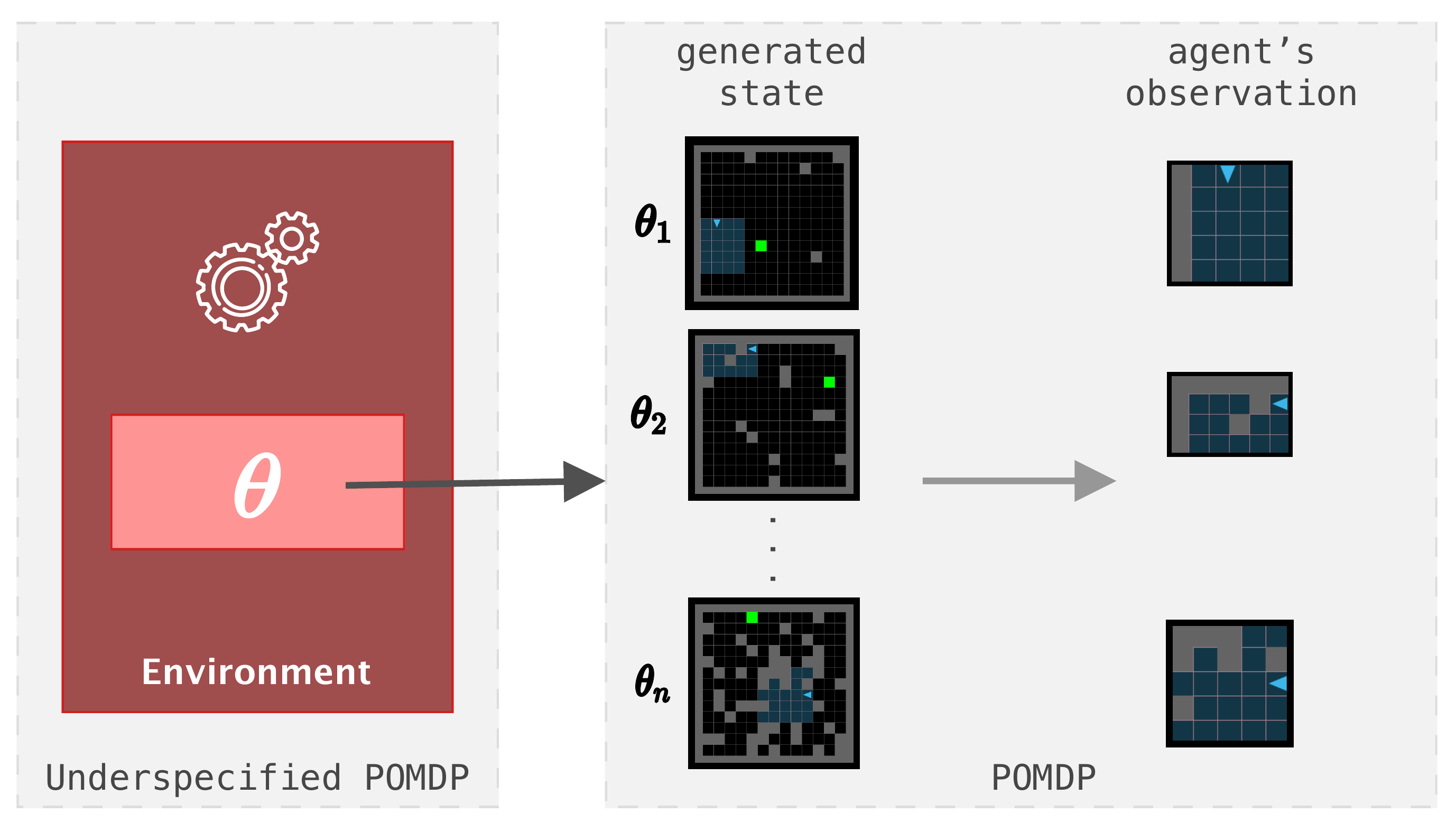}} 
\centering\subfigure[UED Framework]{\includegraphics[width=.61\linewidth]{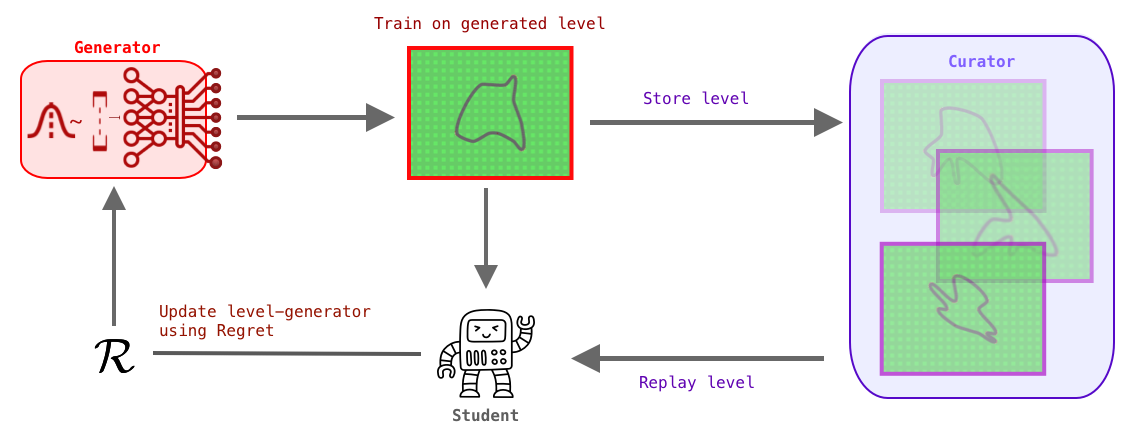}} 
\caption{\small{\textbf{The Unsupervised Environment Design (UED) framework \citep{PAIRED}} 
In an Underspecified POMDP, the layout and dynamics of the environment are left underspecified as a free parameter $\theta$.  The level-generator (either learned or search-based) can then set this parameter to design POMDP levels. (b) This work focuses on the Unsupervised Environment Design (UED) paradigm, which is a method for designing environments using a curriculum, either via a learned adversary to generate new levels or via smart curation of previously-seen levels.}}
\label{fig:ued_framework_fig}
\end{figure}

\section{Background}

\subsection{Reinforcement Learning}

In reinforcement learning (\citep[RL;][]{Sutton1998}, a Markov Decision Process (MDP) is a mathematical framework that models an agent's decision-making process. An MDP is defined as a tuple $(\mathcal{S}, \mathcal{A}, \mathcal{T}, \mathcal{R}, \gamma)$, where: $\mathcal{S}$ is a set of states, $\mathcal{A}$ is a set of actions, $\mathcal{T}$ is the state transition probability function, denoted as $P(s',s|a)$, $\mathcal{R}$ is the reward function, denoted as $R(s,a)$, and $\gamma$ is the discount factor, a scalar between 0 and 1, used to balance the trade-off between immediate and future rewards. The agent's objective in an MDP is to learn an optimal policy, denoted as $\pi$ (which is a mapping from states to actions) that maximizes the expected cumulative discounted reward over time, defined as $\mathbb{E}[\sum_{t=0}^{\infty}\gamma^tR(s_{t},\pi(s_{t}))]$. The value function, denoted as $V^{\pi}(s)$, is a measure of the long-term expected cumulative discounted reward for a given state s, when following a specific policy $\pi$. It is defined as:
$V^{\pi}(s) = \mathbb{E}[\sum_{t=0}^{\infty}\gamma^tR(s_{t},\pi(s_{t})) |  s_0 = s]$.

However, in many environments, the full context of a state is not known to the agent. Rather, the agent gets a partial-view of the state, also called as \textit{observation}, which it then processes for decision-making. Such an MDP is termed as Partially-Observable Markov Decision Process (POMDP). In this case, the tuple is $(\mathcal{S}, \mathcal{A},\mathcal{O},\mathcal{I}, \mathcal{T}, \mathcal{R}, \gamma)$, where $\mathcal{I}: \mathcal{S} \rightarrow \Delta(\mathcal{O})$ is the set of observations that the agent receives. We refer the readers to Figure \ref{fig:ued_framework_fig} (a) for a qualitative representation of POMDPs.

\subsection{Unsupervised Environment Design}
\label{ued_background}

This work considers a more general problem setting where we seek to train an agent across an entire distribution of POMDPs. In this case there are set of design parameters $\mathcal{\theta}$ which control the initial state, transition function, and reward function, e.g. the positions of obstacles in a maze. Given such parameters $\theta$ we can model the target domain as an Underspecified Partially-Observable Markov Decision Process \citep[UPOMDP;][]{PAIRED}. A UPOMDP is defined as a tuple $\mathcal{M} = (\mathcal{\theta},\mathcal{S}, \tilde{\mathcal{S}}_0 \mathcal{A},\mathcal{O},\mathcal{I}, \mathcal{T}, \mathcal{R}, \gamma)$, where  $\mathcal{\theta}$ as the set of free parameters of the environment,
$\mathcal{S}$ is the set of states, 
$\tilde{\mathcal{S}}_0: \mathcal{\theta} \rightarrow \mathcal{S}$ is the initial distribution of states which can depend on the parameters $\theta$, $\mathcal{A}$ as the set of actions, $\mathcal{O}$ as the set of observations, $\mathcal{I}: \mathcal{S} \rightarrow \Delta(\mathcal{O})$ is the observation function, $\mathcal{T}:\mathcal{S} \times \mathcal{A} \times \mathcal{\theta} \rightarrow \Delta(S)$ is the transition function which can depend on the parameters $\theta$, $\mathcal{R}:\mathcal{S} \times \mathcal{\theta} \rightarrow \mathbb{R}$ is the reward function which can depend on the parameters $\theta$, and $\gamma$ is the discount factor.  Here $\Delta(X)$ represents the set of distributions over the set $X$.  Without setting the parameters $\theta$, the environment, and thus the task, is underspecified and the UPOMDP can be thought of as a level editor.  Once the parameters $\theta$ have been specified, the initial state distribution, reward function, and transition function are all fully defined, and the goal of the agent is to maximize the expected discounted sum of rewards as usual (Figure \ref{fig:ued_framework_fig}). 

Unsupervised Environment Design \citep{PAIRED} considers the problem whereby a \emph{teacher} (or adversary) must select levels from this distribution, such that the resulting \emph{student} agent is capable of systematic generalization across all conceivable (solvable) levels. In order to do this, the teacher must maximize a utility function $U$. The most prominent recent approach to UED is to use a teacher that maximizes the concept of \emph{regret}, defined as the difference between the expected return of the current policy $\pi$ and the optimal policy $\pi^*$:

\begin{align}
U_t^R(\pi, \theta) & =\max_{\pi^* \in \Pi}{(\textsc{Regret}^{\theta}(\pi^*,\pi))} \\
& = \max_{\pi^* \in \Pi}{(V^\theta(\pi^*)-V^\theta(\pi))}
\end{align}

Regret-based objectives are desirable as it can be argued, under certain set of assumptions \citep[Theorem 1;][]{PAIRED}, that they promote the creation of the simplest possible levels that the agent cannot currently solve. Moreover, if $S_t= \Pi$ is the set of strategies that the student agent can take and $S_t = \Theta$ is the set of strategies that the teacher can take, then if the learning process reaches a Nash equilibrium, the resulting agent policy $\pi$ provably returns to a minimax regret policy \citep[Theorem 2;][]{PAIRED}, defined as:

\begin{equation}
\pi = \argmin_{\pi_A \in \Pi}{\max_{\theta,\pi_B \in \Theta , \Pi}{(\textsc{Regret}^{\theta}(\pi_A,\pi_B))}}.
\end{equation}

\subsubsection{UED with a Learned Adversary}
\label{PAIRED_background}


\begin{wrapfigure}{R}{0.5\textwidth}
\vspace{-8mm}
    \begin{minipage}{0.5\textwidth}
\begin{algorithm}[H]
\caption{PAIRED \citep{PAIRED}}
\label{alg:PAIRED}

\begin{algorithmic}[1]

\STATE {\bfseries Input:} Initial policies for protagonist $\pi^P$, antagonist $\pi^A$, initial environment generator $\tilde{\Lambda}$
\WHILE{not converged}
\STATE Use $\tilde{\Lambda}$ to generate environment parameters $\vec{\theta}$
\STATE Collect a trajectory $\tau^P$ using $\pi^P$ in environment $\vec{\theta}$
\STATE Update $\pi^P$ to minimize $\mathcal{L}_{ppo}(\pi_P)$
\STATE Collect a trajectory $\tau^A$ using $\pi^A$ in environment $\vec{\theta}$
\STATE Update $\pi^A$ to minimize $\mathcal{L}_{ppo}(\pi_A)$
\STATE Compute the regret as: \\ 
$\normalfont\textsc{Regret}^{\vec{\theta}}(\pi^P, \pi^A) = V^{\vec{\theta}}(\pi^A) - V^{\vec{\theta}}(\pi^P) $ 
\STATE Update $\tilde{\Lambda}$ to maximize regret
\ENDWHILE
\end{algorithmic}
\end{algorithm}
\end{minipage}
\vspace{-8mm}
\end{wrapfigure}

\textit{Protagonist Antagonist Induced Regret Environment Design}  \citep[PAIRED;][]{PAIRED} is a method for generating an adaptive curriculum of levels by training the teacher agent $\pi_{\theta}$ to generate levels which maximize the \emph{relative regret} between \emph{two} student agents, referred to as the protagonist $\pi_p$ and antagonist $\pi_a$.  By designing levels on which the antagonist succeeds and the protagonist fails, the teacher (adversary) in PAIRED develops challenging levels while avoiding the failure case of generating overly difficult or unsolvable levels.  Both the protagonist and antagonist agents are then trained on the generated levels resulting in an emergent curriculum of increasingly complex levels as the teacher adapts to find new challenges the protagonist cannot yet solve. The teacher agent is tasked with maximizing the relative regret between the two student agents using $\normalfont\textsc{Regret}^{\vec{\theta}}(\pi^P, \pi^A) = V^{\vec{\theta}}(\pi^A) - V^{\vec{\theta}}(\pi^P) $. This approach allows for the endless generation of new levels, with the ultimate goal of continually challenging and improving the abilities of the two student agents.

The PAIRED algorithm (shown in Algorithm~\ref{alg:PAIRED}) is an iterative process, where in each iteration, the teacher (adversary) generates the parameters of the environment, $\vec{\theta} \sim \tilde{\Lambda}$, both student agents will play. The protagonist and antagonist agents then generate several trajectories within that same environment. The protagonist is trained to minimize regret, while the antagonist and the teacher (environment adversary) are trained to maximize regret. For more details on the level-generation process as well as the architectures of teacher and student agents, we refer readers to Section \ref{sec:env_details}.

\subsection{UED with Search-based Curator}
\label{sec:ued_curation}

Another approach to UED is Prioritized Level Replay~\citep[PLR;][]{plr, jiang2021robustplr}. This method trains the agent on challenging levels found by curating a rolling buffer of the highest-regret levels surfaced through random search over possible level configurations. PLR approximates regret using the positive value loss, given by:
\begin{equation}
\label{eqn:pvl}
\frac{1}{T}\sum_{t=0}^{T} \max \left(\sum_{k=t}^T(\gamma\lambda)^{k-t}\delta_k, 0\right)
\end{equation}

where $\lambda$ and $\gamma$ are the Generalized Advantage Estimation (GAE) and MDP discount factors respectively, and $\delta_t$, the TD-error at timestep $t$. PLR has been shown to produce policies with strong generalization capabilities, but it is limited to only curating randomly sampled levels. More recently, ACCEL \citep{accel2022} built on PLR to include a mutation operator, making it possible to increase the regret of existing levels. This small modification allows for a rapid escalation in complexity, and results in strong zero-shot transfer generalization, including the challenging BipedalWalker environment. 

Despite these recent innovations, both ACCEL and PLR rely on random samples for level-generation --- and subsequently mutate them randomly. In our work we seek to improve upon PAIRED such that it is competitive with these newer approaches.

\section{Limitations of PAIRED}
\label{limitations}

In theory, if the PAIRED algorithm reaches a Nash equilibrium, the protagonist policy has desirable properties like being at least as good as the antagonist policy on every level. However, in practical implementations, the teacher is trained using RL, which poses a series of challenges. Concretely, the teacher MDP consists of state $s$ which is the current environment configuration and actions $a$ that incrementally change it, until the end of the episode. The teacher receives a sparse reward $r$ once both the student policies have been evaluated in the level.   

This is a difficult RL problem---the teacher must grapple with challenging credit assignment from only getting reward after building a complete level, non-stationarity from updating students and finally a potentially high dimensional problem as there are often many parameters that need to be set in a given environment. For instance, the maze environment has more than 169 parameters which need to be set before the environment designer can receive any reward signal, presenting the environment designer with a high-dimensional exploration problem. 

\subsection{P1: Entropy Collapse}
Indeed, a typical instantiation of PPO struggles to properly explore the space, with the entropy of the policy often collapsing on an overly narrow set of possible level designs. In Section~\ref{sec:entropy_bonus}, we propose to circumvent this challenge by simply adding an entropy bonus which promotes exploration and also acts as a regularizer preventing premature convergence to suboptimal policies. Further, we also explore alternative optimizers in Section~\ref{sec:optimizer} since RL may not be best fit for this blackbox optimization problem. 

\begin{figure}[h!]
        \centering
        \includegraphics[width=0.95\linewidth]{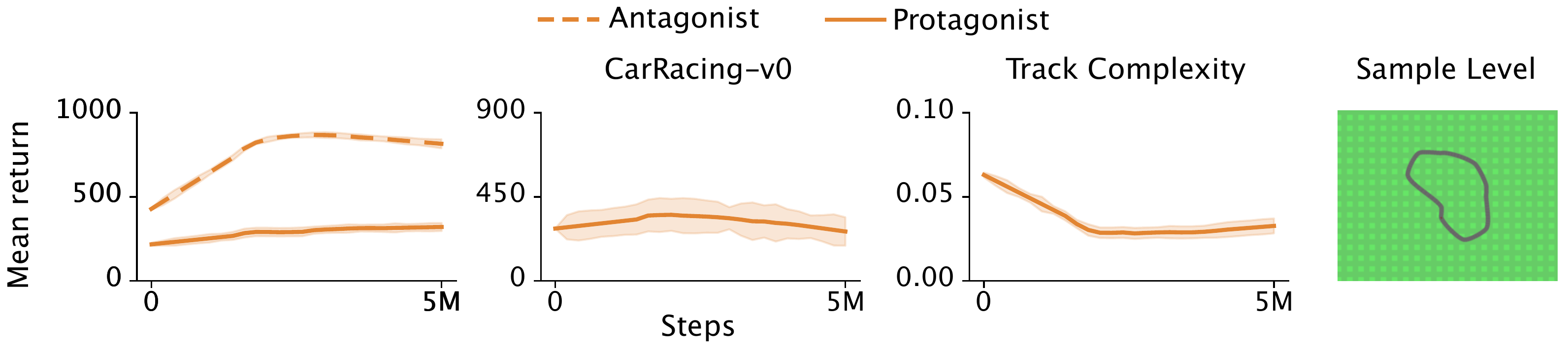}
\caption{\small{\textbf{PAIRED Breaks Down in CarRacing:} \textbf{Left:} Mean returns of protagonist and antagonist on the training levels generated by the adversary. \textbf{Center:} Performance of the protagonist in the CarRacing-v0 validation track as the training progresses. \textbf{Right:} Track complexity of the levels generated by the adversary throughout training. After a while, the training returns of the protagonist saturate, whereas the antagonist keeps on doing well even in very simple tracks (as track complexity decreases severely after ~2M steps of training), thus leading to a never-ending cycle of degenerate levels.}}
\vspace{-5mm}
\label{fig:cr_return_entropy_complexity}
\end{figure}

\subsection{P2: Fall Behind}
As a result of this complexity, the system often converges to suboptimal solutions. For example, in the CarRacing UED environment proposed in \citet{jiang2021robustplr}, $\theta$ corresponds to the coordinates for 12 control points defining a track. As we see in Figure~\ref{fig:cr_return_entropy_complexity}, the adversary finds tracks that initially overly exploit the protagonist agent, which never subsequently recovers. 
Throughout training, the mean return of protagonist always stays lower than that of antagonist, thus leading to a vicious cycle where neither the tracks become complex nor the protagonist is able to learn any useful skills. After 5M steps of training, the protagonist still cannot solve simple round tracks, that are easy for the antagonist agent training with the same algorithm on the same data. This example also demonstrates the significance of getting the right curriculum, with agents unable to learn anything at all after seeing the wrong sequence of initial tasks. 

This example failure mode of PAIRED can also offer a potential solution---in such cases the protagonist's exploration problem can be avoided by treating the antagonist, which is already solving the task by the PAIRED construction, as an \emph{expert demonstrator}. In Section \ref{sec:behavioral_cloning} we describe another possible improvement to PAIRED, adding a behavioral cloning loss to the protagonist so it may learn from the antagonist's solution, inspired by \citet{openai2021asymmetric} in the context of asymmetric self play~\citep{sukhbaatar2017intrinsic}.


\section{Revisiting Design Choices in PAIRED}

In light of the limitations outlined in the previous section, we proposed several design modifications to improve PAIRED. These include the addition of a high entropy bonus (\textbf{HiEnt}), an investigation of the impact of using different optimizers for the level-generating network (\textbf{Evo}), and the implementation of an online behavioral cloning term (\textbf{BiBC, UniBC}). The effectiveness of these modifications will be discussed in the following sections.

Through our experiments, we show that our proposed changes produce more robust and transferable policies. In the following subsections, we will show how our method proposed modifications improves zero-shot generalization over RobustPLR in CarRacing, reaches a performance comparable to ACCEL in Minigrid, and is able to generate complex levels in BipedalWalker by changing the optimizer of the teacher.

\subsection{Entropy Bonus}
\label{sec:entropy_bonus}
In the context of maze environments, empirically we observe during training that the entropy of the teacher (adversary) collapses before training is complete (see Figures \ref{fig:mg25_train_return} and \ref{fig:mg60_train_return}). Therefore, to help alleviate that issue (Problem P1 in Section \ref{introduction}), we propose to add a high entropy bonus to the policy of the agents (and term the baseline as \textbf{PAIRED+HiEnt (HiEnt)}), which encourages both, the teacher and the students (protagonist and antagonist), to explore more robustly. Overall, adding such regularization presents a promising approach for addressing the exploration challenge for both, level-design and action-space. 


\subsection{Optimization Algorithm}
\label{sec:optimizer}

It is common for Deep RL algorithms to utilize Proximal Policy Optimization \citep[PPO;][]{schulman2017proximal} as the objective. However, we hypothesize that the problems P1 and P2 mentioned in Section \ref{limitations} are also affected by the choice of optimizer used inside the teacher agent. In Section \ref{sec:results}, we examine the impact of using different optimizers, specifically a PPO-based teacher (learned editor from PAIRED) and a search-based editor (from ACCEL), on the performance of the level-design process. While a learned editor uses a neural network to design the best level based on the current capabilities of both the student agents, a search-based editor seeks to curate the best level from a pool of randomly generated levels based on a score metric \citep{plr, jiang2021robustplr, accel2022}, and hence the editor here has no learned weights. Our baseline, referred to as \textbf{PAIRED+Evo (Evo)}, is described in the accompanying algorithm \ref{alg:PAIRED_randomoptim} which is based on the algorithm from \cite{accel2022} and we highlight our modifications in violet font. More specifically, we replace the teacher from PAIRED with a random level editor from ACCEL and use the approximate regret (calculated from the antagonist and the protagonist) as the scoring function for replaying or editing the next set of levels.

\subsection{Protagonist-Antagonist Behavioral Cloning}
\label{sec:behavioral_cloning}

\begin{figure*}[h]
\vspace{-3mm}
    \centering
    \begin{minipage}{0.99\textwidth}
    \centering\subfigure{\includegraphics[width=.95\linewidth]{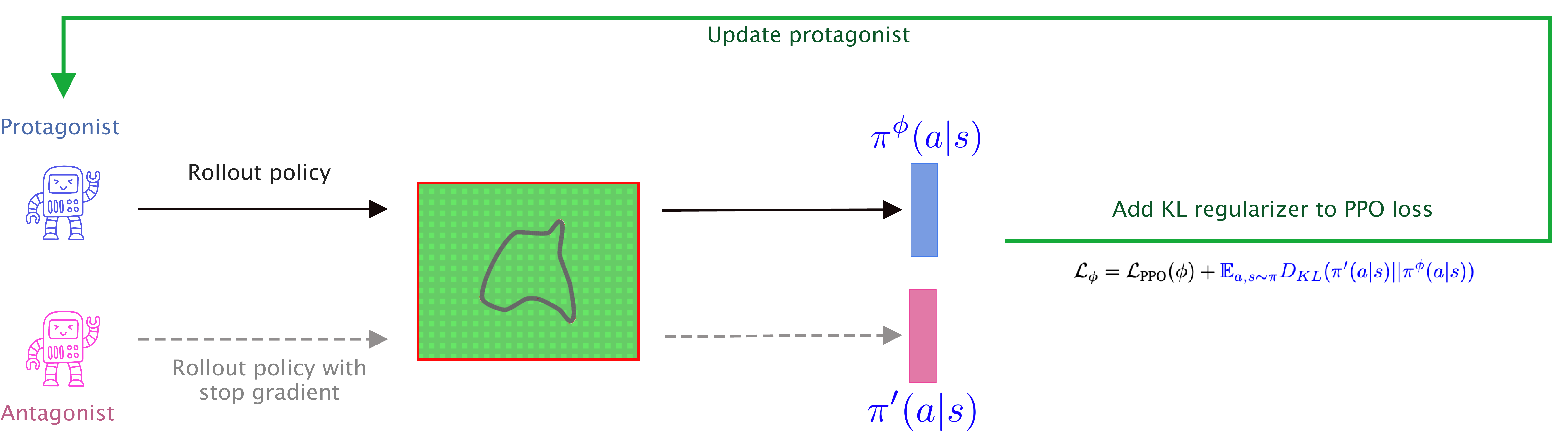}} 
    \vspace{-3mm}
    \caption{PAIRED+BC baseline in the context of the DCD framework \citep{jiang2021robustplr}. PAIRED+BC makes use of an additional regularization term which is added to the PPO loss of both (or either) of the student agents at regular intervals. For more details, refer to Section \ref{sec:behavioral_cloning}. 
    }
    \label{figure:PAIRED_bc_dcd_diagram}
    \end{minipage}
\end{figure*}

The population-based \emph{Peer-to-Peer Online Distillation Strategy for RL} \citep[P2PDRL;][]{zhao2021robust} was demonstrated to improve robustness in a randomized continuous control domain. Taking inspiration from P2PDRL, we propose \textbf{PAIRED+BC}, an online-policy distillation method that generalizes prior methods, whereby the protagonist agent optimizes the RL objective while simultaneously behaviorally cloning the antagonist agent (unidirectional behavioral cloning \textbf{UniBC}), and, if needed, the antagonist agent does the same to the protagonist (bidirectional behavioral cloning \textbf{BiBC}). 

Policy distillation for policy-gradient RL commonly occurs through minimizing the KL divergence of the student's policy $\pi$ from the teacher's policy $\pi'$, $D_{KL}(\pi' || \pi)$. When performed online with PPO training, policy distillation for a student $\pi$ with parameters $\phi$ thus adds an additional regularization term to the PPO objective, $J_{\text{PPO}}(\phi)$:

\begin{equation}
\label{eq:online_policy_distillation_objective}
\mathcal{L}_{\phi} = \mathcal{L}_{\text{PPO}}(\phi) + \textcolor{blue}{\mathbb{E}_{a,s\sim\pi^{\phi}}D_{KL}(\pi'(a|s)||\pi^{\phi}(a|s))},
\end{equation}

PPO performs minibatch SGD over multiple epochs of the collected transition data. To ensure each policy update is regularized by the policy distillation loss, we jointly optimize Equation \ref{eq:online_policy_distillation_objective} periodically in each minibatch update. Algorithm \ref{alg:PAIRED_bc} provides a high-level overview of the PAIRED+BC algorithm.

It should be noted that our best results are achieved with our PAIRED+BC method when utilizing a high entropy bonus for both the students and teacher policies (i.e. UniBC+HiEnt or BiBC+HiEnt). While either of the two baselines, i.e. PAIRED+BC (UniBC or BiBC) without entropy or simply PAIRED+HiEnt, may yield improvements over the PAIRED method alone, they are not sufficient as standalone, optimal approaches. 

\section{Experimental Results}
\label{sec:results}

In this section, we examine the empirical performance of our proposed fixes on three challenging environments, CarRacing F1 \citep{Gym}, Minigrid \citep{gym_minigrid} mazes with 0-60 uniformly sampled blocks \citep{jiang2021robustplr, accel2022}, and the BipedalWalker environment \citep{poet}. We describe the technical details for each of these environments in Appendix \ref{sec:env_details}. For demonstrating the effectiveness of the proposed design choices in highly challenging environments, we also report results on the Minigrid Maze environment with a strict budget of 25-blocks. In this environment, the agents are thus trained on highly sparse grids (see Figure \ref{fig:mg25_tracks}) and tested on the same mazes as in Minigrid 0-60 uniform environment, therefore increasing the difficulty of zero-shot transfer evaluation. All in all, we run the following baselines and ablations in each environment:
\begin{description}
    \item[PAIRED] The original PAIRED algorithm from an open source library\footnote{\href{https://github.com/facebookresearch/dcd}{https://github.com/facebookresearch/dcd}}
    \item[HiEnt] Using the PAIRED baseline from above, and adding high entropy bonuses to the teacher and students
    \item[{Uni/Bi}BC] We run both Bidirection (\textbf{BiBC}) and Unidirectional (\textbf{UniBC}) to demonstrate the effectiveness of both approaches
    \item[{Uni/Bi}BC+HiEnt] To take things further, we combine the two design choices to see how much more we can benefit
    \item[PLR$^\bot$, ACCEL] Additionally, we report the performance of existing SoTA approaches in each environment and also show how our suggested improvements help PAIRED achieve comparable performance with respect to replay-guided UED approaches.
\end{description}

\subsection{CarRacing}

\begin{figure}[h!]
    \vspace{-7mm}
    \centering\subfigure[Tracks]{\includegraphics[width=.45\linewidth]{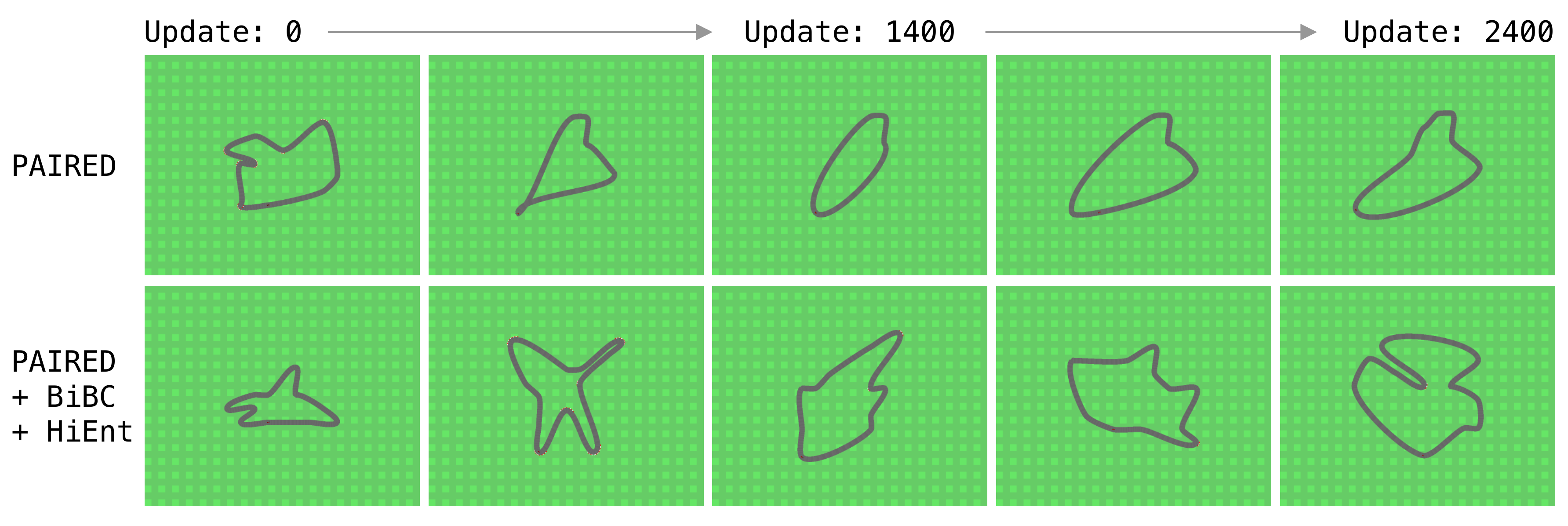}} 
    \hspace{7mm}
\centering\subfigure[Mean Performance]{\includegraphics[width=.45\linewidth]{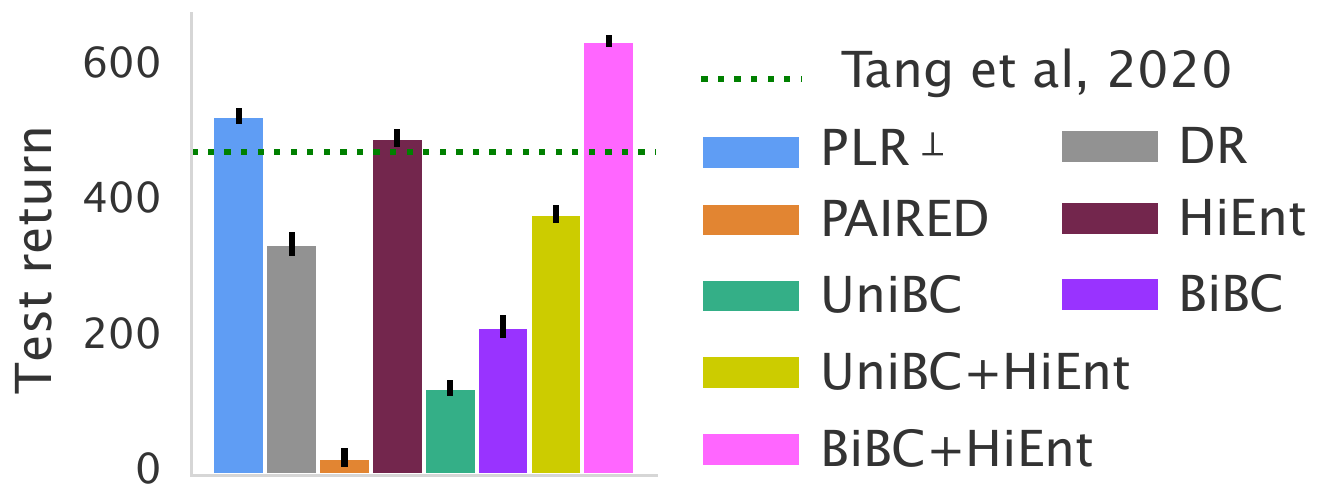}} 
    \vspace{-3mm}
    \caption{(a) F1 levels generated by the adversary as training progresses. (b) Mean aggregate results on real-world F1 tracks (see Figure~\ref{fig:cr_test_tracks}) averaged over 10 seeds (error bars represent the standard deviation). When equipped with a Bidirectional BC term and a high entropy bonus, the protagonist trained in the CarRacing F1 environment (PAIRED+BiBC+HiEnt) is able to generalize better and the teacher agent produces more challenging tracks.}
    \label{fig:cr_f1_mean_aggregate}
\end{figure}

First, we consider the CarRacing environment, previously described in Section~\ref{limitations}. In Figure \ref{fig:cr_f1_mean_aggregate} we show the performance of all PAIRED variants as well as the DR and Robust PLR baselines in the held-out F1 benchmark. As we can see, adding an entropy bonus to the students and the teacher (adversary) is sufficient to avoid the sub-optimal local optimum we previously saw in Figure~\ref{fig:cr_return_entropy_complexity}. Overall, the track complexity increases significantly, which can be seen in Figure~\ref{fig:cr_train_return} and Figure~\ref{fig:cr_f1_mean_aggregate}(a). Furthermore, PAIRED+BiBC+HiEnt stabilizes the open-ended learning process by ensuring that the protagonist is able to learn well from the curriculum (see Figure~\ref{fig:cr_train_return} and provides significant gains over PAIRED, even beating the current SoTA, Robust PLR \citep{jiang2021robustplr} and a non-UED approach \citep{tang2020neuroevolution}, by a clear margin in the CarRacing F1 benchmark. Even the BiBC, UniBC as well as UniBC+HiEnt variants are able to improve the performance of the protagonist over vanilla PAIRED, however, they are not sufficient in making the student agents more competitive.


As well as improving transfer performance, we note the complexity of emergent problems generated by the algorithm drastically improved---tracks generated by the teacher become much more difficult, comprising of sharper edges and turns, compared to those under PAIRED, and start to resemble real-world tracks (see Figure \ref{fig:cr_tracks} in Appendix \ref{sec:additional_results}). 

\subsection{MiniGrid 25-Blocks}

\begin{wrapfigure}{r}{0.53\textwidth}
\vspace{-7mm}
    \includegraphics[width=.53\textwidth]{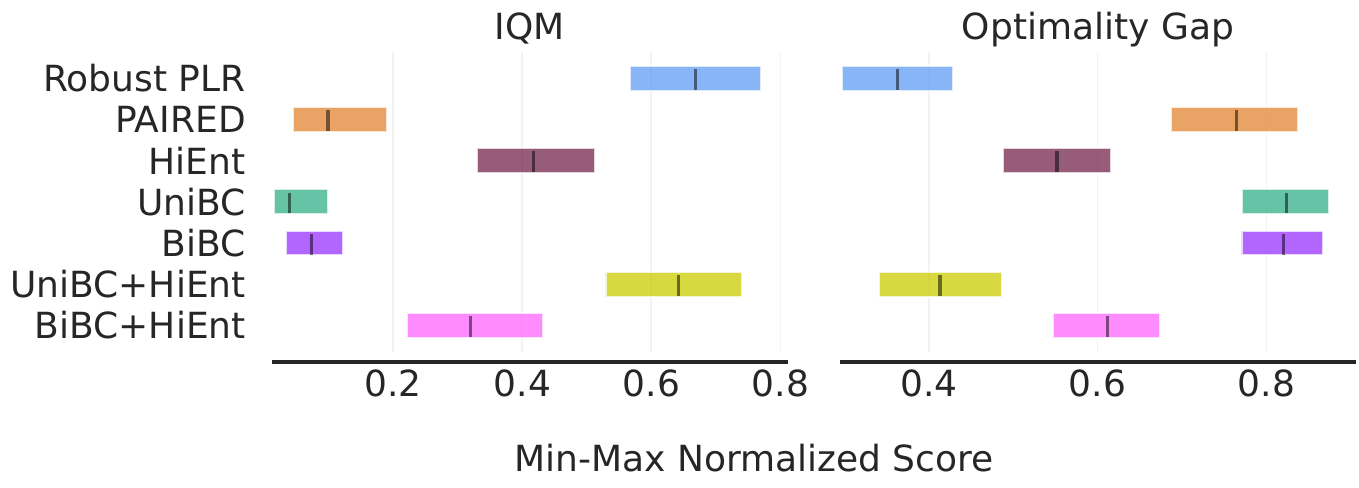}
    \vspace{-7mm}
\caption{\small{IQM and Optimality Gap for all baselines trained on the \textbf{Minigrid 25-Blocks} environment average across 5 seeds. PAIRED+UniBC+HiEnt is able to significantly improve the zero-shot transfer performance over PAIRED when evaluated on 12 test mazes (Table \ref{table:mg25_mean}).}}
   \vspace{-5mm}
\label{fig:mg25_iqm}
\end{wrapfigure}

Next we consider the Minigrid environment which was first introduced in \citet{gym_minigrid}. Readers may refer to Section \ref{sec:env_details} for a summary of this environment. In this experiment, we used the default hyperparameters from \cite{jiang2021robustplr} for PAIRED, which has an entropy coefficient of 0.0 for both, students and teacher. As shown in Figure \ref{fig:mg25_iqm}, simply increasing the entropy bonus coefficient for both the students and the teacher (adversary) leads to significant improvement on the IQM metric (calculated using RLiable library \citep{agarwal2021deep}), and the teacher (adversary) continues generating challenging mazes, as demonstrated in Figure \ref{fig:mg25_train_return}. Since in this environment, there's a sharp drop in the PAIRED's adversary's entropy (see Figure \ref{fig:mg25_train_return}), hence, similar to CarRacing, we found that increasing the entropy bonus helps in improving the performance (see UniBC+HiEnt and BiBC+HiEnt).

Table \ref{table:mg25_mean} shows the performance of the design choices in individual test mazes. We find that in this training environment (with a fixed budget of 25 blocks), the protagonist benefits only by unidirectional cloning (unlike bidirectional BC in the case of CarRacing). This maybe happening because there is very less difference between the training returns of the protagonist and antagonist (Figure \ref{fig:mg25_train_return}) and both the agents are doing equally good in the sparse mazes, thereby eliminating the need for the antagonist to copy protagonist's actions. Overall, PAIRED+UniBC+HiEnt is able to achieve a performance on-par with Robust PLR, thus demonstrating the benefits of our proposed design modifications.

\begin{table}[h]
\centering
\vspace{-3mm}
\caption{\small{Zero-shot test performance results for each baseline in 12 challenging minigrid mazes when trained with a budget of \textbf{25-blocks} only. Each baseline has been evaluated for 5 seeds and 100 episodes per seed. We report mean and standard deviation for Solved Rate as the metric here, highlighting in bold the best performing agents.}}
    
\resizebox{0.95\columnwidth}{!}{%
\begin{tabular}{lllllllr}
\toprule
Environment &Robust PLR &PAIRED &HiEnt &UniBC &BiBC &UniBC+HiEnt &BiBC+HiEnt\\
\midrule
Labyrinth&$\mathbf{0.66\pm0.16}$&$0.27\pm0.17$&$0.28\pm0.18$&$0.0\pm0.0$&$0.0\pm0.0$&$\mathbf{0.77\pm0.15}$&$0.19\pm0.13$\\
Labyrinth2&$\mathbf{0.56\pm0.13}$&$\mathbf{0.32\pm0.2}$&$0.24\pm0.18$&$0.0\pm0.0$&$0.0\pm0.0$&$\mathbf{0.41\pm0.18}$&$0.21\pm0.14$\\
LargeCorridor&$\mathbf{0.71\pm0.16}$&$0.26\pm0.17$&$\mathbf{0.68\pm0.18}$&$0.38\pm0.22$&$0.48\pm0.17$&$\mathbf{0.59\pm0.19}$&$\mathbf{0.79\pm0.14}$\\
Maze&$\mathbf{0.4\pm0.18}$&$0.0\pm0.0$&$0.11\pm0.09$&$0.0\pm0.0$&$0.0\pm0.0$&$\mathbf{0.26\pm0.19}$&$0.07\pm0.07$\\
Maze2&$\mathbf{0.65\pm0.16}$&$0.0\pm0.0$&$0.11\pm0.08$&$0.0\pm0.0$&$0.0\pm0.0$&$\mathbf{0.6\pm0.15}$&$0.33\pm0.14$\\
Maze3&$\mathbf{0.67\pm0.16}$&$0.2\pm0.2$&$\mathbf{0.69\pm0.16}$&$0.0\pm0.0$&$0.03\pm0.03$&$\mathbf{0.58\pm0.15}$&$\mathbf{0.38\pm0.23}$\\
MiniGrid-FourRooms&$\mathbf{0.51\pm0.04}$&$0.33\pm0.04$&$\mathbf{0.45\pm0.02}$&$0.42\pm0.04$&$0.35\pm0.02$&$\mathbf{0.53\pm0.06}$&$0.42\pm0.02$\\
MiniGrid-SimpleCrossingS11N5&$\mathbf{0.91\pm0.04}$&$0.3\pm0.14$&$0.73\pm0.04$&$0.36\pm0.08$&$0.37\pm0.1$&$0.74\pm0.1$&$\mathbf{0.76\pm0.12}$\\
PerfectMazeMedium&$\mathbf{0.44\pm0.12}$&$0.17\pm0.09$&$\mathbf{0.31\pm0.04}$&$0.04\pm0.02$&$0.09\pm0.03$&$\mathbf{0.42\pm0.05}$&$0.24\pm0.06$\\
SixteenRooms&$\mathbf{0.87\pm0.05}$&$0.46\pm0.19$&$\mathbf{0.68\pm0.14}$&$0.31\pm0.18$&$0.19\pm0.16$&$\mathbf{0.79\pm0.09}$&$0.14\pm0.08$\\
SixteenRoomsFewerDoors&$0.48\pm0.17$&$0.18\pm0.16$&$0.4\pm0.05$&$0.1\pm0.09$&$0.18\pm0.12$&$\mathbf{0.75\pm0.09}$&$0.27\pm0.17$\\
SmallCorridor&$\mathbf{0.78\pm0.09}$&$0.35\pm0.19$&$\mathbf{0.67\pm0.17}$&$0.51\pm0.18$&$0.48\pm0.17$&$\mathbf{0.59\pm0.21}$&$\mathbf{0.86\pm0.08}$\\
\midrule
Mean&$\mathbf{0.64\pm0.03}$&$0.24\pm0.09$&$0.45\pm0.03$&$0.18\pm0.03$&$0.18\pm0.05$&$\mathbf{0.59\pm0.06}$&$0.39\pm0.05$\\
\bottomrule
\end{tabular}
}

\vspace{-3mm}
\label{table:mg25_mean}
\end{table}

\subsection{MiniGrid 0-60 Uniform-Blocks}

\begin{figure}[H]
    \centering\subfigure[IQM]{\includegraphics[width=.45\linewidth]{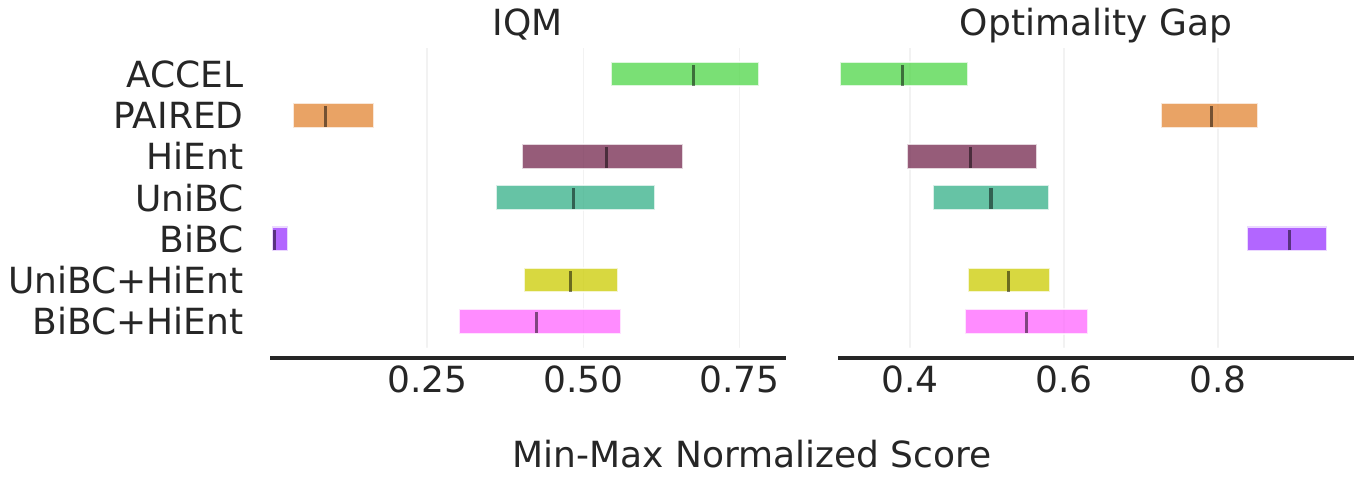}} 
\centering\subfigure[Maze Complexity metrics]{\includegraphics[width=.43\linewidth]{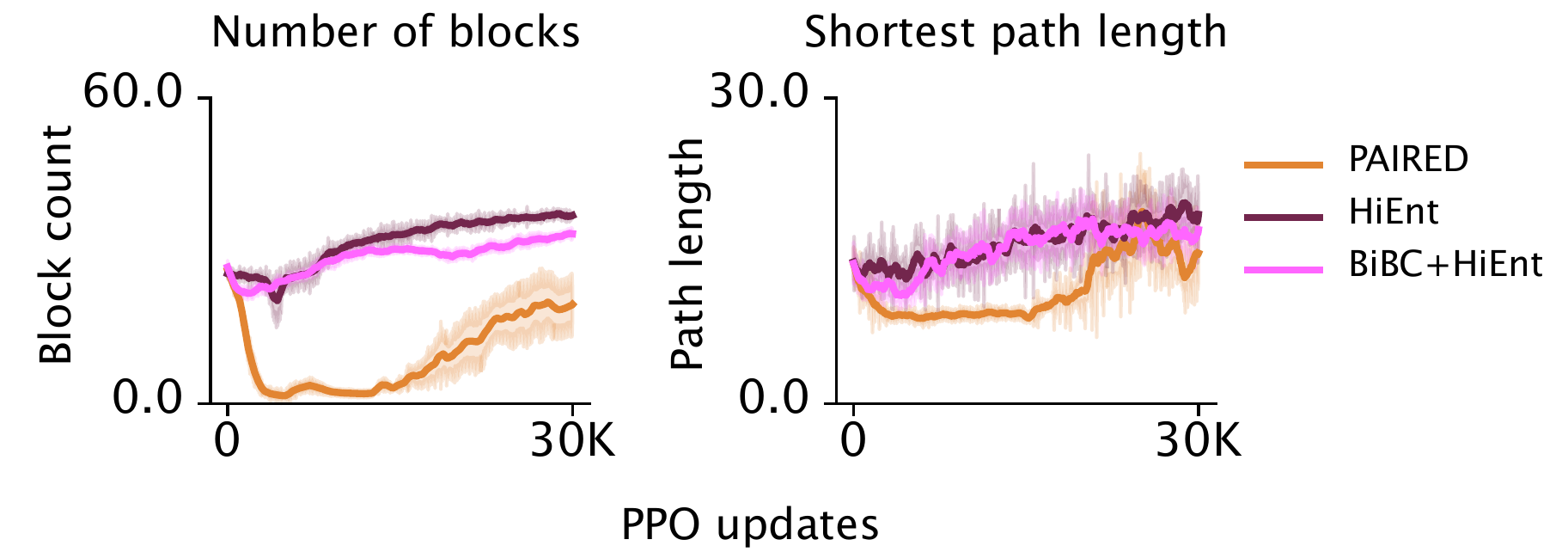}} 
\centering\subfigure[Tracks]{\includegraphics[width=.88\linewidth]{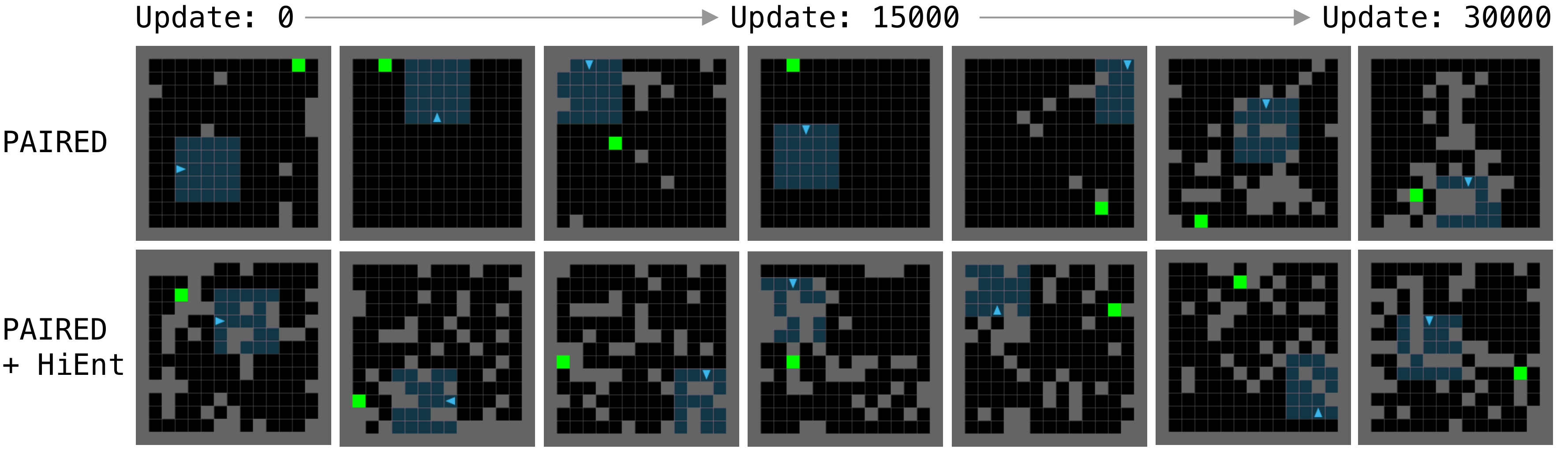}} 
    \vspace{-3mm}
    \caption{\small{{\textbf{Minigrid 0-60 Uniform-Blocks environment.} (a) IQM and Optimality Gap for all baselines when evaluated on test mazes (Table \ref{table:mg60_mean}) averaged over 5 seeds. (b) Maze complexity metrics showing the emergent complexity during training in HiEnt and BiBC+HiEnt methods. (c) Sample tracks generated by the teacher agent (adversary) in PAIRED and HiEnt methods. Due to entropy collapse (Problem P1), PAIRED's adversary generates sparse mazes leading to poor zero-shot transfer at test time, and adding a high entropy bonus alleviates that issue.}}}
    \label{fig:mg60_mean_aggregate}
\end{figure}

Figure \ref{fig:mg60_mean_aggregate} (a) shows the performance of the HiEnt baseline versus the original PAIRED method on the Minigrid 0-60 Uniform Blocks benchmark. The HiEnt baseline consists of 0.05 entropy coefficient for the level-generating teacher (adversary), and a 0.005 bonus for the students. Moreover, Figure \ref{fig:mg60_mean_aggregate} (b) shows the number of blocks per level and the minimum path length to reach the goal (with a length of 0 assigned to unsolvable levels). These figures illustrate that the grids generated by PAIRED's adversary kept on becoming sparse as training progressed, thus leading to a degraded performance. Both HiEnt and BiBC+HiEnt baselines improve on the complexity front and we found that the performance of the HiEnt baseline can significantly reduce the gap with ACCEL, often matching performance on highly challenging test mazes (see Table~\ref{table:mg60_mean}). Readers may note that we train all PAIRED baselines for 30k gradient updates (250M environment steps), whereas we train ACCEL for 20k gradient updates only (which corresponds to ~400M environment steps).
However, in Minigrid 0-60 (unlike Minigrid 25-Blocks benchmark), the level-generating process doesn't benefit the most from behavioral cloning, as the HiEnt baseline achieves the highest gains instead of BiBC+HiEnt or UniBC+HiEnt. For a comparison of all these methods on individual test environments, readers may refer to Table \ref{table:mg60_mean}.

\subsection{Changing the optimizer in BipedalWalker and Minigrid}
\label{sec:accel_paired}
\begin{figure}[H]
    \vspace{-5mm}
    \centering\subfigure[Test IQM]{\includegraphics[width=.4\linewidth]{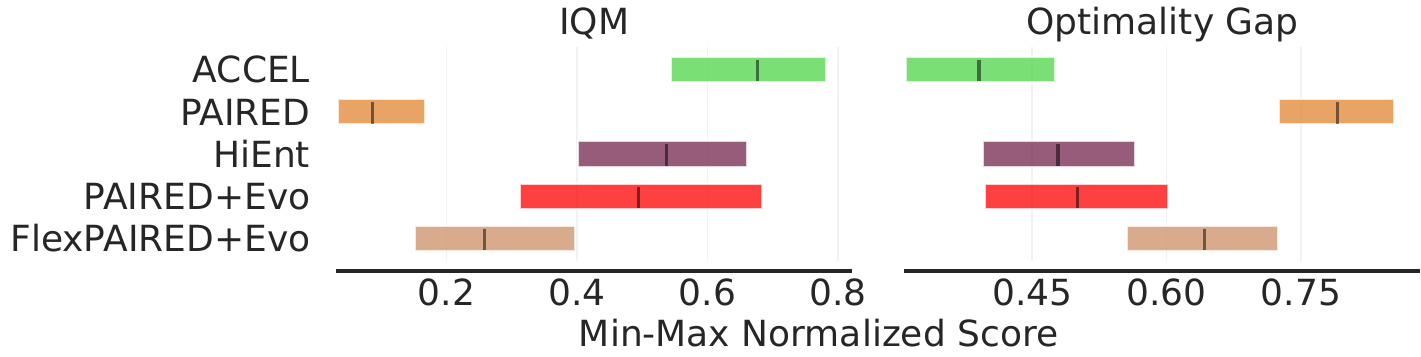}} 
\centering\subfigure[Validation Solved Rates]{\includegraphics[width=.55\linewidth]{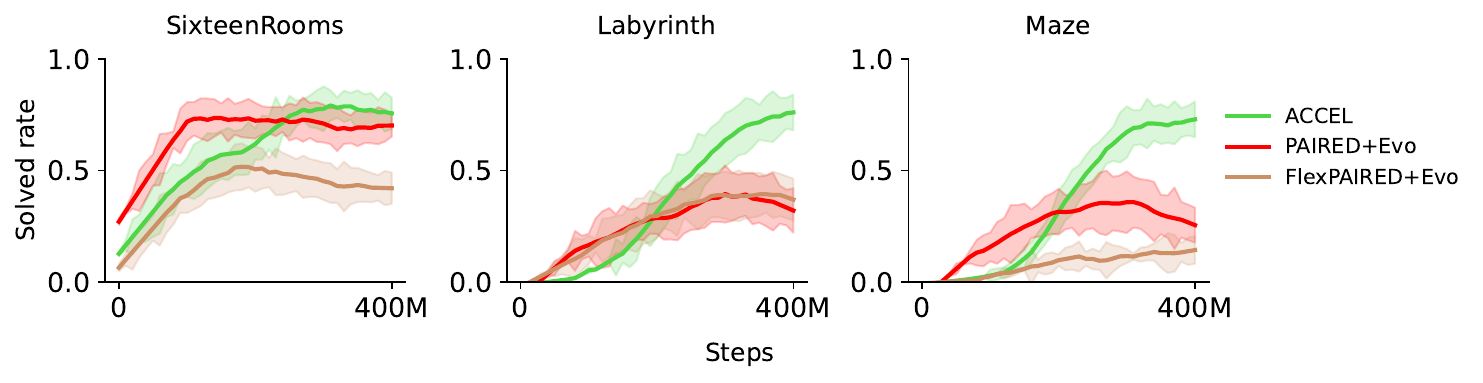}} 
    \vspace{-3mm}
    \caption{IQM, Optimality Gap and Solved Rates averaged across 5 seeds for PAIRED+Evo and FlexPAIRED+Evo baselines on Minigrid 0-60 Uniform-blocks environment.}
    \label{fig:mg60_optim}
    \vspace{-3mm}
\end{figure}

\begin{wrapfigure}{l}{0.5\textwidth}
    \vspace{-3mm}
    \includegraphics[width=0.5\textwidth, height=4.5cm, clip]{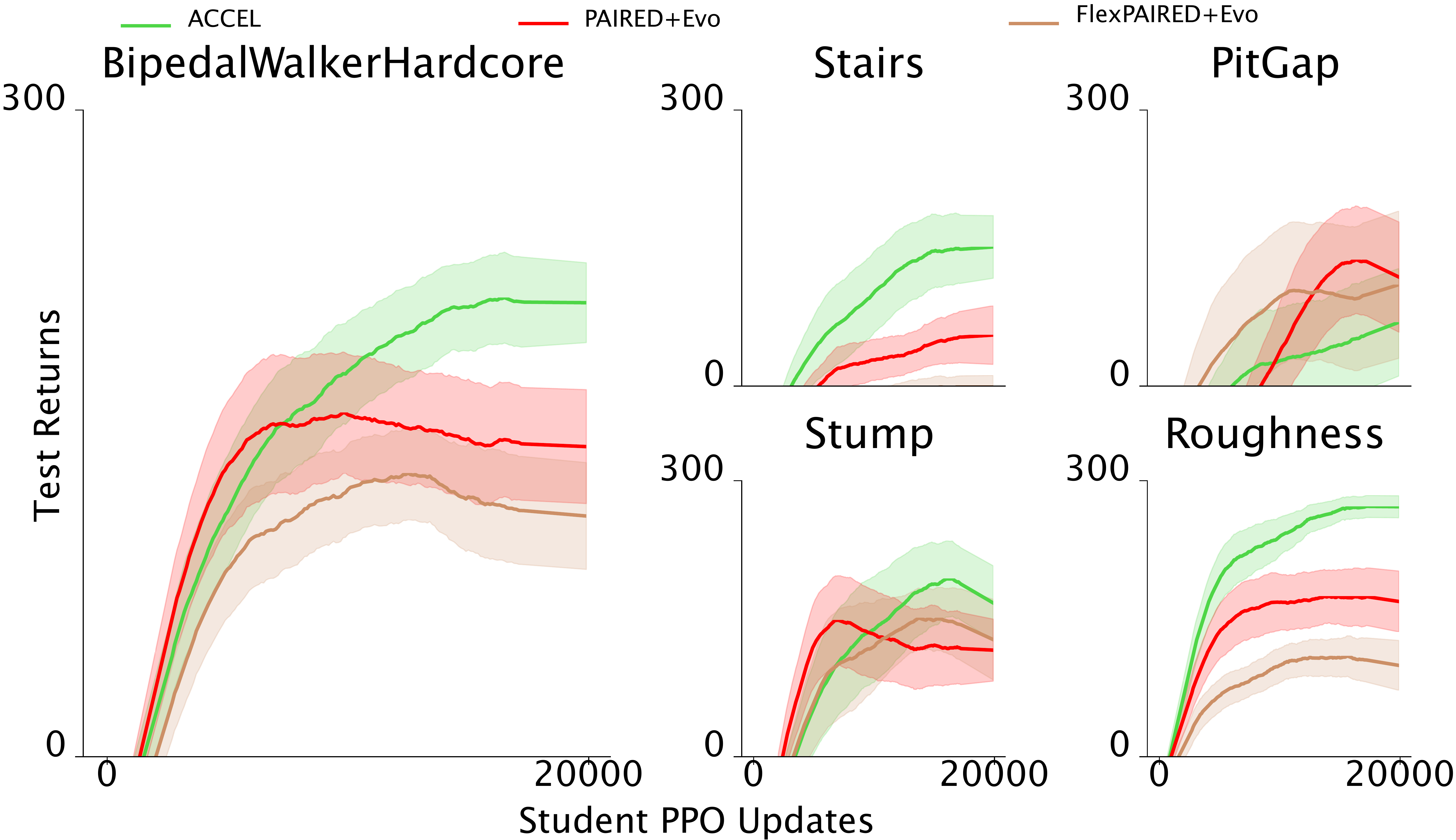}
    \vspace{-6mm}
    \caption{\small{Zero-shot transfer performance of PAIRED+Evo on 5 test environments as training progresses in the BipedalWalker environment.}}
    \label{fig:bpw_train}
    \vspace{-3mm}
\end{wrapfigure}

In this section, we study the effect of changing the optimizer from a learned teacher to a random search-based level-designer, naming the baseline to be PAIRED+Evo. Additionally, we also report the performance of this baseline when using a \emph{flexible} regret objective \citep{PAIRED, flexpaired} where there is no fixed antagonist or protagonist. We call this baseline FlexPAIRED+Evo.

As depicted in Figure~\ref{fig:mg60_optim}(a), both PAIRED+Evo and FlexPAIRED+Evo underperform compared to PAIRED+HiEnt in the Minigrid 0-60 Uniform Blocks environment, suggesting that a high entropy bonus has a greater impact than the choice of optimizer. Moreover, in Figure \ref{fig:mg60_optim}(b), it can be seen that the ACCEL tends to do a lot better on the three validation mazes, indicating that the positive value loss leads to better curation of training levels compared to the approximate regret calculated by a pair of student agents in PAIRED(or Flexible PAIRED)+Evo. Similar to Minigrid, in the challenging BipedalWalker \citep{poet} environment, PAIRED+Evo performs worse than ACCEL, but better than PAIRED (Figure~\ref{fig:bpw_train}), indicating that the choice of optimizer is a significant design factor and warrants further investigation, especially in more challenging environments where the level-design space is even bigger. 

Note that while each of these design choices has been tested in isolation, it would be particularly interesting to see how PAIRED performs when combined with all three design choices, i.e. PAIRED+Evo+BC+HiEnt. The goal of our work was to rigorously test individual design choices rather than produce a single state-of-the-art agent, thus, we run a small experiment in Appendix~\ref{app:acp_all} where we compare a combination of these design choices without any hyperparameter tuning.


\section{Related Work}
Our work extends a popular line of research in regret-based Unsupervised Environment Design (UED), which aims to address problems of generalization and robustness in RL.

\textbf{Unsupervised Environment Design}
Unsupervised Environment Design (UED), which automatically designs training environments to maximize learning potential is a quickly growing field
\citep{poet, pinsky, enhanced_poet, PAIRED, xland, jiang2021robustplr, accel2022, jiang2022grounding, dharna2022transfer, team2023human}.  A significant line of this work is targeted at designing high-regret environments \citep{PAIRED, flexpaired, jiang2021robustplr, accel2022, ada}. Many of the most recent successful techniques have used curation or evolution to build individual levels \citep{jiang2021robustplr, accel2022, team2023human}.  While these approaches are currently state of the art they build each level one by one.  Thus a promising direction UED is for neural models to generate environment parameters, thus allowing for combinatorial generalization across levels.  

PAIRED~\citep{PAIRED} and the algorithms which are built on it takes this neural generative approach \citep{flexpaired, du2022takes, jiang2021robustplr, wang2022zone}.  Thus our work gives guidance to many approaches on how they may be tuned to be competitive with the state of the art. Unsupervised environment design is also very closely related to work on AI for procedural environment generation. AI for procedural environment generation is a long-standing
~\citep{togelius2011search, browne2010evolutionary, togelius2008evolving} and active field~\citep{earle2021illuminating, khalifa2022mutation, dsage}.  Though our work focuses on the case of using the resulting environments to train an agent and thus does not directly address these approaches, cross-pollinating the ideas of these fields is an important line of future work.

\textbf{Generalization in RL}
In addition to providing a curriculum, regret-based UED approaches have been shown to improve robustness and generalization in RL \citep{kirk2021generalisation}.  Deep RL systems, like the deep neural networks on which they are based~\citep{szegedy2013intriguing}, have been shown to exhibit robustness failures such as failing under adversarial attacks
\citep{kos2017delving, lin2017tactics, gleave2019adversarial} or overfitting to the environment configuration~\citep{witty2021measuring, di2022goal}.  Many approaches have aimed to remedy these failures by changing the objective of RL systems, aiming to solve something closer to a robust MDP~\citep{bagnell2001solving,iyengar2005robust,nilim2005robust}.  There have been a wide variety of such approaches
\citep{garcia2015comprehensive}, including those which aim to solve regret-based MDP formulations \citep{ghavamzadeh2016safe, regan2011robust, regan2012regret}.  

A powerful and scalable approach for solving robust MDPs is adversarial training, for which there exist many specialized methods
\citep{pinto2017robust, pinto2017advrobotics, morimoto2005robust}.  The regret-based UED algorithms we study in this work can be seen as a high dimensional adversarial training approach to a particular sort of solve regret-based MDP.


\section{Conclusion}
This work considers problems arising when using a learned adversary (or teacher) for unsupervised environment design (UED). Focusing on the popular PAIRED algorithm, we discussed potential instabilities during training and explored several approaches for mitigating them. Concretely, our results indicate that properly tuning the entropy parameter is crucial for achieving good performance---by doing so, we find that PAIRED is able to achieve results that are competitive with other state-of-the-art UED methods. We also consider the use of behavioral cloning, which helps the student agent in learning useful skills in the simplest of levels. However, in some environments where this is not an issue, behavioral cloning can lead to reduced out-of-distribution (OOD) robustness in the student model. Moreover, we find that, when combined with effective tuning for the entropy coefficient, the choice of optimizer only makes a marginal difference in the performance. 

In general, our results make it possible to achieve state-of-the-art results with a \emph{learned adversary} (or teacher), propelling this class of methods back to the forefront of UED research. We believe this could result in a variety of future innovation, such as combining our proposed solutions with population-based training.



\subsubsection*{Acknowledgments}
We would like to thank Roberta Raileanu, Mikayel Samvelyan, Eric Hambro, and Jakob Foerster for providing valuable feedback and engaging in insightful discussions about our work. We are also grateful to the anonymous reviewers for their constructive suggestions. This work was funded by Meta AI.



\bibliography{collas2023_conference}

\begin{thebibliography}{55}
\providecommand{\natexlab}[1]{#1}
\providecommand{\url}[1]{\texttt{#1}}
\expandafter\ifx\csname urlstyle\endcsname\relax
  \providecommand{\doi}[1]{doi: #1}\else
  \providecommand{\doi}{doi: \begingroup \urlstyle{rm}\Url}\fi

\bibitem[{Adaptive Agent Team} et~al.(2023){Adaptive Agent Team}, Bauer,
  Baumli, Baveja, Behbahani, Bhoopchand, Bradley-Schmieg, Chang, Clay,
  Collister, Dasagi, Gonzalez, Gregor, Hughes, Kashem, Loks-Thompson, Openshaw,
  Parker-Holder, Pathak, Perez-Nieves, Rakicevic, Rocktäschel, Schroecker,
  Sygnowski, Tuyls, York, Zacherl, and Zhang]{ada}
{Adaptive Agent Team}, Jakob Bauer, Kate Baumli, Satinder Baveja, Feryal
  Behbahani, Avishkar Bhoopchand, Nathalie Bradley-Schmieg, Michael Chang,
  Natalie Clay, Adrian Collister, Vibhavari Dasagi, Lucy Gonzalez, Karol
  Gregor, Edward Hughes, Sheleem Kashem, Maria Loks-Thompson, Hannah Openshaw,
  Jack Parker-Holder, Shreya Pathak, Nicolas Perez-Nieves, Nemanja Rakicevic,
  Tim Rocktäschel, Yannick Schroecker, Jakub Sygnowski, Karl Tuyls, Sarah
  York, Alexander Zacherl, and Lei Zhang.
\newblock Human-timescale adaptation in an open-ended task space, 2023.
\newblock URL \url{https://arxiv.org/abs/2301.07608}.

\bibitem[Agarwal et~al.(2021)Agarwal, Schwarzer, Castro, Courville, and
  Bellemare]{agarwal2021deep}
Rishabh Agarwal, Max Schwarzer, Pablo~Samuel Castro, Aaron Courville, and
  Marc~G. Bellemare.
\newblock Deep reinforcement learning at the edge of the statistical precipice.
\newblock In \emph{Advances in Neural Information Processing Systems}. 2021.

\bibitem[Bagnell et~al.(2001)Bagnell, Ng, and Schneider]{bagnell2001solving}
J~Andrew Bagnell, Andrew~Y Ng, and Jeff~G Schneider.
\newblock Solving uncertain markov decision processes.
\newblock 2001.

\bibitem[Berner et~al.(2019)Berner, Brockman, Chan, Cheung, Debiak, Dennison,
  Farhi, Fischer, Hashme, Hesse, J{\'{o}}zefowicz, Gray, Olsson, Pachocki,
  Petrov, de~Oliveira~Pinto, Raiman, Salimans, Schlatter, Schneider, Sidor,
  Sutskever, Tang, Wolski, and Zhang]{dota}
Christopher Berner, Greg Brockman, Brooke Chan, Vicki Cheung, Przemyslaw
  Debiak, Christy Dennison, David Farhi, Quirin Fischer, Shariq Hashme, Chris
  Hesse, Rafal J{\'{o}}zefowicz, Scott Gray, Catherine Olsson, Jakub Pachocki,
  Michael Petrov, Henrique~Pond{\'{e}} de~Oliveira~Pinto, Jonathan Raiman, Tim
  Salimans, Jeremy Schlatter, Jonas Schneider, Szymon Sidor, Ilya Sutskever,
  Jie Tang, Filip Wolski, and Susan Zhang.
\newblock Dota 2 with large scale deep reinforcement learning.
\newblock \emph{CoRR}, abs/1912.06680, 2019.

\bibitem[Bhatt et~al.(2022)Bhatt, Tjanaka, Fontaine, and Nikolaidis]{dsage}
Varun Bhatt, Bryon Tjanaka, Matthew~C Fontaine, and Stefanos Nikolaidis.
\newblock Deep surrogate assisted generation of environments.
\newblock \emph{arXiv preprint arXiv:2206.04199}, 2022.

\bibitem[Brockman et~al.(2016)Brockman, Cheung, Pettersson, Schneider,
  Schulman, Tang, and Zaremba]{Gym}
Greg Brockman, Vicki Cheung, Ludwig Pettersson, Jonas Schneider, John Schulman,
  Jie Tang, and Wojciech Zaremba.
\newblock Open{AI} {G}ym, 2016.

\bibitem[Browne \& Maire(2010)Browne and Maire]{browne2010evolutionary}
Cameron Browne and Frederic Maire.
\newblock Evolutionary game design.
\newblock \emph{IEEE Transactions on Computational Intelligence and AI in
  Games}, 2\penalty0 (1):\penalty0 1--16, 2010.

\bibitem[Chevalier-Boisvert et~al.(2018)Chevalier-Boisvert, Willems, and
  Pal]{gym_minigrid}
Maxime Chevalier-Boisvert, Lucas Willems, and Suman Pal.
\newblock Minimalistic gridworld environment for {O}pen{AI} {G}ym.
\newblock \url{https://github.com/maximecb/gym-minigrid}, 2018.

\bibitem[Degrave et~al.(2022)Degrave, Felici, Buchli, Neunert, Tracey,
  Carpanese, Ewalds, Hafner, Abdolmaleki, de~Las~Casas,
  et~al.]{degrave2022magnetic}
Jonas Degrave, Federico Felici, Jonas Buchli, Michael Neunert, Brendan Tracey,
  Francesco Carpanese, Timo Ewalds, Roland Hafner, Abbas Abdolmaleki, Diego
  de~Las~Casas, et~al.
\newblock Magnetic control of tokamak plasmas through deep reinforcement
  learning.
\newblock \emph{Nature}, 602\penalty0 (7897):\penalty0 414--419, 2022.

\bibitem[Dennis et~al.(2020)Dennis, Jaques, Vinitsky, Bayen, Russell, Critch,
  and Levine]{PAIRED}
Michael Dennis, Natasha Jaques, Eugene Vinitsky, Alexandre Bayen, Stuart
  Russell, Andrew Critch, and Sergey Levine.
\newblock Emergent complexity and zero-shot transfer via unsupervised
  environment design.
\newblock In \emph{Advances in Neural Information Processing Systems},
  volume~33, 2020.

\bibitem[Dharna et~al.(2020)Dharna, Togelius, and Soros]{pinsky}
Aaron Dharna, Julian Togelius, and L.~B. Soros.
\newblock Co-generation of game levels and game-playing agents.
\newblock \emph{Proceedings of the AAAI Conference on Artificial Intelligence
  and Interactive Digital Entertainment}, 16\penalty0 (1):\penalty0 203--209,
  Oct. 2020.

\bibitem[Dharna et~al.(2022)Dharna, Hoover, Togelius, and
  Soros]{dharna2022transfer}
Aaron Dharna, Amy~K Hoover, Julian Togelius, and Lisa Soros.
\newblock Transfer dynamics in emergent evolutionary curricula.
\newblock \emph{IEEE Transactions on Games}, 2022.

\bibitem[Di~Langosco et~al.(2022)Di~Langosco, Koch, Sharkey, Pfau, and
  Krueger]{di2022goal}
Lauro~Langosco Di~Langosco, Jack Koch, Lee~D Sharkey, Jacob Pfau, and David
  Krueger.
\newblock Goal misgeneralization in deep reinforcement learning.
\newblock In \emph{International Conference on Machine Learning}, pp.\
  12004--12019. PMLR, 2022.

\bibitem[Du et~al.(2022)Du, Abbeel, and Grover]{du2022takes}
Yuqing Du, Pieter Abbeel, and Aditya Grover.
\newblock It takes four to tango: Multiagent selfplay for automatic curriculum
  generation.
\newblock \emph{arXiv preprint arXiv:2202.10608}, 2022.

\bibitem[Earle et~al.(2021)Earle, Snider, Fontaine, Nikolaidis, and
  Togelius]{earle2021illuminating}
Sam Earle, Justin Snider, Matthew~C. Fontaine, Stefanos Nikolaidis, and Julian
  Togelius.
\newblock Illuminating diverse neural cellular automata for level generation,
  2021.

\bibitem[Garc{\i}a \& Fern{\'a}ndez(2015)Garc{\i}a and
  Fern{\'a}ndez]{garcia2015comprehensive}
Javier Garc{\i}a and Fernando Fern{\'a}ndez.
\newblock A comprehensive survey on safe reinforcement learning.
\newblock \emph{Journal of Machine Learning Research}, 16\penalty0
  (1):\penalty0 1437--1480, 2015.

\bibitem[Ghavamzadeh et~al.(2016)Ghavamzadeh, Petrik, and
  Chow]{ghavamzadeh2016safe}
Mohammad Ghavamzadeh, Marek Petrik, and Yinlam Chow.
\newblock Safe policy improvement by minimizing robust baseline regret.
\newblock \emph{Advances in Neural Information Processing Systems}, 29, 2016.

\bibitem[Gleave et~al.(2019)Gleave, Dennis, Wild, Kant, Levine, and
  Russell]{gleave2019adversarial}
Adam Gleave, Michael Dennis, Cody Wild, Neel Kant, Sergey Levine, and Stuart
  Russell.
\newblock Adversarial policies: Attacking deep reinforcement learning.
\newblock \emph{arXiv preprint arXiv:1905.10615}, 2019.

\bibitem[Gur et~al.(2021)Gur, Jaques, Miao, Choi, Tiwari, Lee, and
  Faust]{flexpaired}
Izzeddin Gur, Natasha Jaques, Yingjie Miao, Jongwook Choi, Manoj Tiwari,
  Honglak Lee, and Aleksandra Faust.
\newblock Environment generation for zero-shot compositional reinforcement
  learning.
\newblock \emph{Advances in Neural Information Processing Systems}, 34, 2021.

\bibitem[Hu et~al.(2021)Hu, Lerer, Cui, Pineda, Brown, and Foerster]{hu2021off}
Hengyuan Hu, Adam Lerer, Brandon Cui, Luis Pineda, Noam Brown, and Jakob
  Foerster.
\newblock Off-belief learning.
\newblock In \emph{International Conference on Machine Learning}, pp.\
  4369--4379. PMLR, 2021.

\bibitem[Iyengar(2005)]{iyengar2005robust}
Garud~N Iyengar.
\newblock Robust dynamic programming.
\newblock \emph{Mathematics of Operations Research}, 30\penalty0 (2):\penalty0
  257--280, 2005.

\bibitem[Jiang et~al.(2021{\natexlab{a}})Jiang, Dennis, Parker{-}Holder,
  Foerster, Grefenstette, and Rockt{\"{a}}schel]{jiang2021robustplr}
Minqi Jiang, Michael Dennis, Jack Parker{-}Holder, Jakob Foerster, Edward
  Grefenstette, and Tim Rockt{\"{a}}schel.
\newblock Replay-guided adversarial environment design.
\newblock In \emph{Advances in Neural Information Processing Systems}.
  2021{\natexlab{a}}.

\bibitem[Jiang et~al.(2021{\natexlab{b}})Jiang, Grefenstette, and
  Rockt{\"{a}}schel]{plr}
Minqi Jiang, Edward Grefenstette, and Tim Rockt{\"{a}}schel.
\newblock Prioritized level replay.
\newblock In \emph{The International Conference on Machine Learning}.
  2021{\natexlab{b}}.

\bibitem[Jiang et~al.(2022)Jiang, Dennis, Parker-Holder, Lupu, K{\"u}ttler,
  Grefenstette, Rockt{\"a}schel, and Foerster]{jiang2022grounding}
Minqi Jiang, Michael Dennis, Jack Parker-Holder, Andrei Lupu, Heinrich
  K{\"u}ttler, Edward Grefenstette, Tim Rockt{\"a}schel, and Jakob Foerster.
\newblock Grounding aleatoric uncertainty in unsupervised environment design.
\newblock \emph{arXiv preprint arXiv:2207.05219}, 2022.

\bibitem[Khalifa et~al.(2022)Khalifa, Togelius, and Green]{khalifa2022mutation}
Ahmed Khalifa, Julian Togelius, and Michael~Cerny Green.
\newblock Mutation models: Learning to generate levels by imitating evolution.
\newblock In \emph{Proceedings of the 17th International Conference on the
  Foundations of Digital Games}, pp.\  1--9, 2022.

\bibitem[Kirk et~al.(2021)Kirk, Zhang, Grefenstette, and
  Rockt{\"{a}}schel]{kirk2021generalisation}
Robert Kirk, Amy Zhang, Edward Grefenstette, and Tim Rockt{\"{a}}schel.
\newblock A survey of generalisation in deep reinforcement learning.
\newblock \emph{CoRR}, abs/2111.09794, 2021.

\bibitem[Kos \& Song(2017)Kos and Song]{kos2017delving}
Jernej Kos and Dawn Song.
\newblock Delving into adversarial attacks on deep policies.
\newblock \emph{arXiv preprint arXiv:1705.06452}, 2017.

\bibitem[Lin et~al.(2017)Lin, Hong, Liao, Shih, Liu, and Sun]{lin2017tactics}
Yen-Chen Lin, Zhang-Wei Hong, Yuan-Hong Liao, Meng-Li Shih, Ming-Yu Liu, and
  Min Sun.
\newblock Tactics of adversarial attack on deep reinforcement learning agents.
\newblock \emph{arXiv preprint arXiv:1703.06748}, 2017.

\bibitem[Morimoto \& Doya(2005)Morimoto and Doya]{morimoto2005robust}
Jun Morimoto and Kenji Doya.
\newblock Robust reinforcement learning.
\newblock \emph{Neural computation}, 17\penalty0 (2):\penalty0 335--359, 2005.

\bibitem[Nilim \& El~Ghaoui(2005)Nilim and El~Ghaoui]{nilim2005robust}
Arnab Nilim and Laurent El~Ghaoui.
\newblock Robust control of markov decision processes with uncertain transition
  matrices.
\newblock \emph{Operations Research}, 53\penalty0 (5):\penalty0 780--798, 2005.

\bibitem[{Open Ended Learning Team} et~al.(2021){Open Ended Learning Team},
  Stooke, Mahajan, Barros, Deck, Bauer, Sygnowski, Trebacz, Jaderberg, Mathieu,
  McAleese, Bradley{-}Schmieg, Wong, Porcel, Raileanu, Hughes{-}Fitt, Dalibard,
  and Czarnecki]{xland}
{Open Ended Learning Team}, Adam Stooke, Anuj Mahajan, Catarina Barros, Charlie
  Deck, Jakob Bauer, Jakub Sygnowski, Maja Trebacz, Max Jaderberg,
  Micha{\"{e}}l Mathieu, Nat McAleese, Nathalie Bradley{-}Schmieg, Nathaniel
  Wong, Nicolas Porcel, Roberta Raileanu, Steph Hughes{-}Fitt, Valentin
  Dalibard, and Wojciech~Marian Czarnecki.
\newblock Open-ended learning leads to generally capable agents.
\newblock \emph{CoRR}, abs/2107.12808, 2021.

\bibitem[OpenAI et~al.(2021)OpenAI, Plappert, Sampedro, Xu, Akkaya, Kosaraju,
  Welinder, D'Sa, Petron, de~Oliveira~Pinto, Paino, Noh, Weng, Yuan, Chu, and
  Zaremba]{openai2021asymmetric}
OpenAI OpenAI, Matthias Plappert, Raul Sampedro, Tao Xu, Ilge Akkaya, Vineet
  Kosaraju, Peter Welinder, Ruben D'Sa, Arthur Petron, Henrique~Ponde
  de~Oliveira~Pinto, Alex Paino, Hyeonwoo Noh, Lilian Weng, Qiming Yuan, Casey
  Chu, and Wojciech Zaremba.
\newblock Asymmetric self-play for automatic goal discovery in robotic
  manipulation, 2021.

\bibitem[Parker-Holder et~al.(2022)Parker-Holder, Jiang, Dennis, Samvelyan,
  Foerster, Grefenstette, and Rockt{\"a}schel]{accel2022}
Jack Parker-Holder, Minqi Jiang, Michael Dennis, Mikayel Samvelyan, Jakob
  Foerster, Edward Grefenstette, and Tim Rockt{\"a}schel.
\newblock Evolving curricula with regret-based environment design.
\newblock \emph{arXiv preprint arXiv:2203.01302}, 2022.

\bibitem[Pinto et~al.(2017{\natexlab{a}})Pinto, Davidson, and
  Gupta]{pinto2017advrobotics}
Lerrel Pinto, James Davidson, and Abhinav Gupta.
\newblock Supervision via competition: Robot adversaries for learning tasks.
\newblock In \emph{2017 IEEE International Conference on Robotics and
  Automation (ICRA)}, pp.\  1601--1608, 2017{\natexlab{a}}.
\newblock \doi{10.1109/ICRA.2017.7989190}.

\bibitem[Pinto et~al.(2017{\natexlab{b}})Pinto, Davidson, Sukthankar, and
  Gupta]{pinto2017robust}
Lerrel Pinto, James Davidson, Rahul Sukthankar, and Abhinav Gupta.
\newblock Robust adversarial reinforcement learning.
\newblock In \emph{International Conference on Machine Learning}, pp.\
  2817--2826. PMLR, 2017{\natexlab{b}}.

\bibitem[Regan \& Boutilier(2011)Regan and Boutilier]{regan2011robust}
Kevin Regan and Craig Boutilier.
\newblock Robust online optimization of reward-uncertain mdps.
\newblock In \emph{IJCAI Proceedings-International Joint Conference on
  Artificial Intelligence}, volume~22, pp.\  2165, 2011.

\bibitem[Regan \& Boutilier(2012)Regan and Boutilier]{regan2012regret}
Kevin Regan and Craig Boutilier.
\newblock Regret-based reward elicitation for markov decision processes.
\newblock \emph{arXiv preprint arXiv:1205.2619}, 2012.

\bibitem[Schulman et~al.(2017)Schulman, Wolski, Dhariwal, Radford, and
  Klimov]{schulman2017proximal}
John Schulman, Filip Wolski, Prafulla Dhariwal, Alec Radford, and Oleg Klimov.
\newblock Proximal policy optimization algorithms.
\newblock \emph{arXiv preprint arXiv:1707.06347}, 2017.

\bibitem[Silver et~al.(2016)Silver, Huang, Maddison, Guez, Sifre, van~den
  Driessche, Schrittwieser, Antonoglou, Panneershelvam, Lanctot, Dieleman,
  Grewe, Nham, Kalchbrenner, Sutskever, Lillicrap, Leach, Kavukcuoglu, Graepel,
  and Hassabis]{alphago}
David Silver, Aja Huang, Chris~J. Maddison, Arthur Guez, Laurent Sifre, George
  van~den Driessche, Julian Schrittwieser, Ioannis Antonoglou, Vedavyas
  Panneershelvam, Marc Lanctot, Sander Dieleman, Dominik Grewe, John Nham, Nal
  Kalchbrenner, Ilya Sutskever, Timothy~P. Lillicrap, Madeleine Leach, Koray
  Kavukcuoglu, Thore Graepel, and Demis Hassabis.
\newblock Mastering the game of {G}o with deep neural networks and tree search.
\newblock \emph{Nature}, 529:\penalty0 484--489, 2016.

\bibitem[Silver et~al.(2017)Silver, Hubert, Schrittwieser, Antonoglou, Lai,
  Guez, Lanctot, Sifre, Kumaran, Graepel, et~al.]{alphazero}
David Silver, Thomas Hubert, Julian Schrittwieser, Ioannis Antonoglou, Matthew
  Lai, Arthur Guez, Marc Lanctot, Laurent Sifre, Dharshan Kumaran, Thore
  Graepel, et~al.
\newblock Mastering chess and shogi by self-play with a general reinforcement
  learning algorithm.
\newblock \emph{arXiv preprint arXiv:1712.01815}, 2017.

\bibitem[Song et~al.(2020)Song, Jiang, Tu, Du, and
  Neyshabur]{observational_overfitting}
Xingyou Song, Yiding Jiang, Stephen Tu, Yilun Du, and Behnam Neyshabur.
\newblock Observational overfitting in reinforcement learning.
\newblock In \emph{8th International Conference on Learning Representations,
  {ICLR} 2020, Addis Ababa, Ethiopia, April 26-30, 2020}. OpenReview.net, 2020.

\bibitem[Sukhbaatar et~al.(2017)Sukhbaatar, Lin, Kostrikov, Synnaeve, Szlam,
  and Fergus]{sukhbaatar2017intrinsic}
Sainbayar Sukhbaatar, Zeming Lin, Ilya Kostrikov, Gabriel Synnaeve, Arthur
  Szlam, and Rob Fergus.
\newblock Intrinsic motivation and automatic curricula via asymmetric
  self-play.
\newblock \emph{arXiv preprint arXiv:1703.05407}, 2017.

\bibitem[Sutton \& Barto(1998)Sutton and Barto]{Sutton1998}
Richard~S. Sutton and Andrew~G. Barto.
\newblock \emph{Introduction to Reinforcement Learning}.
\newblock MIT Press, Cambridge, MA, USA, 1st edition, 1998.
\newblock ISBN 0262193981.

\bibitem[Szegedy et~al.(2013)Szegedy, Zaremba, Sutskever, Bruna, Erhan,
  Goodfellow, and Fergus]{szegedy2013intriguing}
Christian Szegedy, Wojciech Zaremba, Ilya Sutskever, Joan Bruna, Dumitru Erhan,
  Ian Goodfellow, and Rob Fergus.
\newblock Intriguing properties of neural networks.
\newblock \emph{arXiv preprint arXiv:1312.6199}, 2013.

\bibitem[Tang et~al.(2020)Tang, Nguyen, and Ha]{tang2020neuroevolution}
Yujin Tang, Duong Nguyen, and David Ha.
\newblock Neuroevolution of self-interpretable agents.
\newblock In \emph{Proceedings of the 2020 Genetic and Evolutionary Computation
  Conference}, pp.\  414--424, 2020.

\bibitem[Team et~al.(2023)Team, Bauer, Baumli, Baveja, Behbahani, Bhoopchand,
  Bradley-Schmieg, Chang, Clay, Collister, et~al.]{team2023human}
Adaptive~Agent Team, Jakob Bauer, Kate Baumli, Satinder Baveja, Feryal
  Behbahani, Avishkar Bhoopchand, Nathalie Bradley-Schmieg, Michael Chang,
  Natalie Clay, Adrian Collister, et~al.
\newblock Human-timescale adaptation in an open-ended task space.
\newblock \emph{arXiv preprint arXiv:2301.07608}, 2023.

\bibitem[Togelius \& Schmidhuber(2008)Togelius and
  Schmidhuber]{togelius2008evolving}
Julian Togelius and Jurgen Schmidhuber.
\newblock An experiment in automatic game design.
\newblock In \emph{2008 IEEE Symposium On Computational Intelligence and
  Games}, pp.\  111--118, 2008.
\newblock \doi{10.1109/CIG.2008.5035629}.

\bibitem[Togelius et~al.(2011)Togelius, Yannakakis, Stanley, and
  Browne]{togelius2011search}
Julian Togelius, Georgios~N Yannakakis, Kenneth~O Stanley, and Cameron Browne.
\newblock Search-based procedural content generation: A taxonomy and survey.
\newblock \emph{IEEE Transactions on Computational Intelligence and AI in
  Games}, 3\penalty0 (3):\penalty0 172--186, 2011.

\bibitem[Vinyals et~al.(2019)Vinyals, Babuschkin, Czarnecki, Mathieu, Dudzik,
  Chung, Choi, Powell, Ewalds, Georgiev, Oh, Horgan, Kroiss, Danihelka, Huang,
  Sifre, Cai, Agapiou, Jaderberg, Vezhnevets, Leblond, Pohlen, Dalibard,
  Budden, Sulsky, Molloy, Paine, G{\"{u}}l{\c{c}}ehre, Wang, Pfaff, Wu, Ring,
  Yogatama, W{\"{u}}nsch, McKinney, Smith, Schaul, Lillicrap, Kavukcuoglu,
  Hassabis, Apps, and Silver]{alphastar}
Oriol Vinyals, Igor Babuschkin, Wojciech~M. Czarnecki, Micha{\"{e}}l Mathieu,
  Andrew Dudzik, Junyoung Chung, David~H. Choi, Richard Powell, Timo Ewalds,
  Petko Georgiev, Junhyuk Oh, Dan Horgan, Manuel Kroiss, Ivo Danihelka, Aja
  Huang, Laurent Sifre, Trevor Cai, John~P. Agapiou, Max Jaderberg,
  Alexander~Sasha Vezhnevets, R{\'{e}}mi Leblond, Tobias Pohlen, Valentin
  Dalibard, David Budden, Yury Sulsky, James Molloy, Tom~L. Paine, {\c{C}}aglar
  G{\"{u}}l{\c{c}}ehre, Ziyu Wang, Tobias Pfaff, Yuhuai Wu, Roman Ring, Dani
  Yogatama, Dario W{\"{u}}nsch, Katrina McKinney, Oliver Smith, Tom Schaul,
  Timothy~P. Lillicrap, Koray Kavukcuoglu, Demis Hassabis, Chris Apps, and
  David Silver.
\newblock Grandmaster level in starcraft {II} using multi-agent reinforcement
  learning.
\newblock \emph{Nat.}, 575\penalty0 (7782):\penalty0 350--354, 2019.
\newblock \doi{10.1038/s41586-019-1724-z}.

\bibitem[Wang et~al.(2022)Wang, Mu, Arumugam, Jaques, and
  Goodman]{wang2022zone}
Rose~E Wang, Jesse Mu, Dilip Arumugam, Natasha Jaques, and Noah Goodman.
\newblock In the zone: Measuring difficulty and progression in curriculum
  generation.
\newblock In \emph{Deep Reinforcement Learning Workshop NeurIPS 2022}, 2022.

\bibitem[Wang et~al.(2019)Wang, Lehman, Clune, and Stanley]{poet}
Rui Wang, Joel Lehman, Jeff Clune, and Kenneth~O. Stanley.
\newblock Paired open-ended trailblazer {(POET):} endlessly generating
  increasingly complex and diverse learning environments and their solutions.
\newblock \emph{CoRR}, abs/1901.01753, 2019.

\bibitem[Wang et~al.(2020)Wang, Lehman, Rawal, Zhi, Li, Clune, and
  Stanley]{enhanced_poet}
Rui Wang, Joel Lehman, Aditya Rawal, Jiale Zhi, Yulun Li, Jeffrey Clune, and
  Kenneth Stanley.
\newblock Enhanced {POET}: Open-ended reinforcement learning through unbounded
  invention of learning challenges and their solutions.
\newblock In Hal~Daumé III and Aarti Singh (eds.), \emph{Proceedings of the
  37th International Conference on Machine Learning}, volume 119 of
  \emph{Proceedings of Machine Learning Research}, pp.\  9940--9951. PMLR,
  13--18 Jul 2020.

\bibitem[Witty et~al.(2021)Witty, Lee, Tosch, Atrey, Clary, Littman, and
  Jensen]{witty2021measuring}
Sam Witty, Jun~K Lee, Emma Tosch, Akanksha Atrey, Kaleigh Clary, Michael~L
  Littman, and David Jensen.
\newblock Measuring and characterizing generalization in deep reinforcement
  learning.
\newblock \emph{Applied AI Letters}, 2\penalty0 (4):\penalty0 e45, 2021.

\bibitem[Zhang et~al.(2018)Zhang, Ballas, and
  Pineau]{zhang2018generalizationcont}
Amy Zhang, Nicolas Ballas, and Joelle Pineau.
\newblock A dissection of overfitting and generalization in continuous
  reinforcement learning.
\newblock \emph{CoRR}, abs/1806.07937, 2018.

\bibitem[Zhao \& Hospedales(2021)Zhao and Hospedales]{zhao2021robust}
Chenyang Zhao and Timothy Hospedales.
\newblock Robust domain randomised reinforcement learning through peer-to-peer
  distillation.
\newblock In \emph{Asian Conference on Machine Learning}, pp.\  1237--1252.
  PMLR, 2021.

\end{thebibliography}
\bibliographystyle{collas2023_conference}

\newpage
\appendix
\section{Experimental Details}
\label{sec:env_details}

In this section, we summarize the technical details of each of the environment used and list out the hyperparameters used for each of the baselines in Table \ref{table:hyperparams}.

\subsection{F1 CarRacing}
\paragraph{\textbf{Environment}} We use the CarRacing F1 UED benchmark introduced in \cite{jiang2021robustplr}. The environment used is a reparameterized version of the CarRacing game \cite{Gym}, where the tracks consist of closed loops made of Bézier curves created from 12 randomly sampled control points. The student agent is rewarded for driving over each unvisited polygon and penalized for each time step. The student agent's observation space consists of a 96x96x3 pixel RGB image with a bird's eye view of the vehicle, which allows the agent to make decisions based on the current state of the environment and the action space is a 3-dimensional continuous action, corresponding to control values for steer, gas and brake. The adversary (teacher agent) generates a sequence of 12 control points, one per time step, within a fixed radius of the center of the playfield, using a 10x10 grid to encode the control points, which is embedded using 2D convolutions and fully connected layers. The protagonists and antagonists train via PPO using 2D CNN operations. We refer the reader to Appendix of \cite{jiang2021robustplr} for more details about the architecture.

\begin{figure}[h]
\centering
\includegraphics[width=0.99\linewidth]{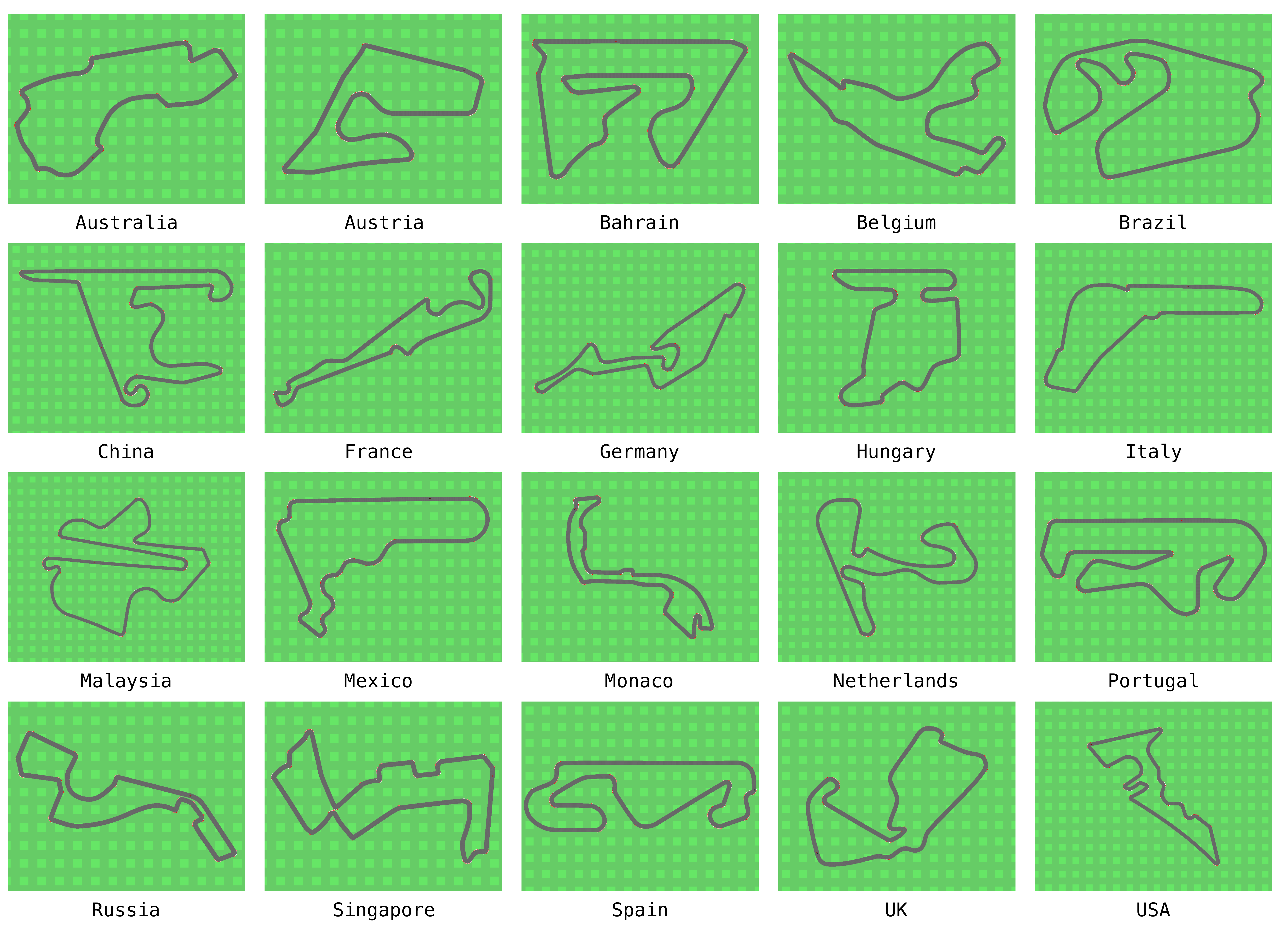}
\caption{Test tracks used for assessing zero-shot performance in CarRacing F1 environment.}
\label{fig:cr_test_tracks}
\end{figure}

\paragraph{\textbf{Hyperparameters}} Since our proposed methodology builds off PAIRED, we inherited most of the parameters from \citet{jiang2021robustplr}. For HiEnt baseline, we swept over adversary and student agent entropies in the range  $\{ \text{0.0, 0.005, 0.05, 0.1}\}$ for the students and $\{ \text{0.0, 0.005, 0.05, 0.1, 0.2}\}$ for the adversary. For the BC baselines (both UniBC and BiBC), we swept over the following hyperparameters:  $\{ \text{0.005, 0.01, 0.05, 0.1, 0.5}\}$ for the KL Loss coefficient, $\{ \text{10, 25}\}$ gradient updates for the KL Loss interval, $\{ \text{0.0, 0.005, 0.05, 0.1}\}$ for the student entropy and $\{ \text{0.0, 0.005, 0.05, 0.1, 0.2}\}$ for the adversary entropy. Similar to \citet{jiang2021robustplr}, we used Test Returns from CarRacingF1-Germany, CarRacingF1-Italy and CarRacingF1-Singapore for validation performance averaged across 3 seeds. Figure \ref{fig:cr_test_tracks} shows all the tracks which were used to assess zero-shot transfer performance at test time for 10 seeds.

\subsection{Minigrid}
\paragraph{\textbf{Environment}}  Using the same setup as in \citep{PAIRED,jiang2021robustplr}, a maze-based game named MiniGrid \cite{gym_minigrid} is used as the environment. The agent's objective is to navigate the maze and reach the end while avoiding walls. The agent receives a reward of $1-\frac{T}{T_{max}}$ if it reaches the end, where $T$ is the episode length and $T_{max}$ is the maximum episode length (set to 250). If the agent fails to reach the end, a reward of 0 is given. The agent's perception of the environment includes its orientation and a 7x7 grid that encompasses the agent and the area immediately in front of it. The agent can take 7 actions, but only 3 of them are used in the game: turn left, turn right, and move forward. Both student agents, the protagonist and the antagonist, follow this. The mazes are generated by an adversary (the teacher agent) which is given $N$ steps to place walls in a 13x13 grid, and then chooses the location of the end and the agent's starting position. The value of $N$ is a constant 25 in the case of 25-blocks fixed budget experiment, and in the case of 0-60 Uniform-blocks, the value of $N$ is uniformly sampled between [0,60]. The generator architecture uses a convolution layer to process the full grid observation, followed by an LSTM with a hidden dimension of 256 and two fully connected layers to produce action logits over the possible 169 cells. The student architecture is similar to the generator architecture, but it uses a convolution with 16 filters to process the partial observation and doesn't use random noise.

\begin{figure}[h]
\centering
\includegraphics[width=0.8\linewidth]{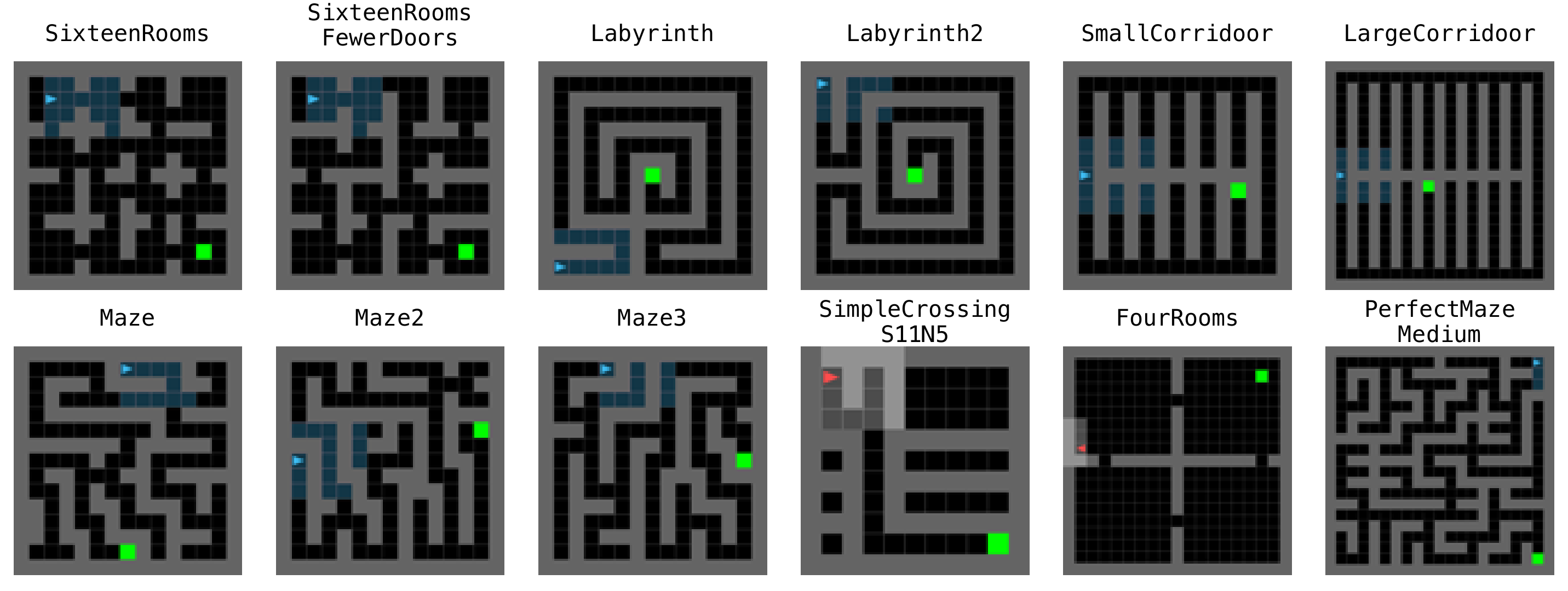}
\caption{Test mazes used for assessing zero-shot performance in the Minigrid environment.}
\label{fig:mg_test_mazes}
\end{figure}

\paragraph{\textbf{Hyperparameters}} Here, for the HiEnt baseline, we swept over adversary and student agent entropies in the range  $\{ \text{0.0, 0.005, 0.05, 0.1}\}$ for the students and $\{ \text{0.0, 0.005, 0.05, 0.1}\}$ for the adversary. For the BC baselines (both UniBC and BiBC), we swept over the following the hyperparameters:  $\{ \text{0.005, 0.01, 0.05, 0.1, 0.5}\}$ for the KL Loss coefficient, $\{ \text{5, 10, 25}\}$ gradient updates for the KL Loss interval, $\{ \text{0.0, 0.005, 0.05, 0.1}\}$ for the student entropy and $\{ \text{0.0, 0.005, 0.05, 0.1}\}$ for the adversary entropy. For the Evo experiments, we use all the hyperparameters same as ACCEL, except the level replay coefficient $\rho$ which we sweep over $\{ \text{0.5, 0.9}\}$.  We use SixteenRooms, Labyrinth and Maze for assessing validation performance by running 3 seeds per experiment, and report the final results on challenging mazes (see Figure \ref{fig:mg_test_mazes}) by averaging over 5 seeds.

\subsection{BipedalWalker}
\paragraph{\textbf{Environment}}  We use the environment introduced in \citet{accel2022} and refer the readers to Appendix C \citep{accel2022} for implementation details of the environment as well as the ACCEL baseline.

\paragraph{\textbf{Hyperparameters}} Our PAIRED+Evo baseline is based off ACCEL, with the addition of one extra agent. Hence, all our hyperparameters are same as that of ACCEL, with the modifications highlighted in Algorithm \ref{alg:PAIRED_randomoptim}.

\begin{table}[H]
\caption{Table summarizing the hyperparameters for each of the methods.}
\label{table:hyperparams}
\vskip 0.15in
\begin{center}
\begin{small}
\begin{sc}
\resizebox{0.85\columnwidth}{!}{%
\begin{tabular}{lcccc}
\toprule
\textbf{Parameter} & CarRacing & MiniGrid 0-60 Uniform & Minigrid 25 & BipedalWalker \\
\midrule

\emph{\textbf{PPO}} & \\
$\gamma$ & 0.99  & 0.995 & 0.995 & 0.99 \\
$\lambda_{\text{GAE}}$ & 0.9  & 0.95 & 0.95 & 0.9 \\
PPO rollout length & 125  & 256 & 256 & 2000 \\
PPO epochs & 8  & 5 & 5 & 5 \\
PPO minibatches per epoch & 4 & 1 & 1 & 32 \\
PPO clip range & 0.2 & 0.2 & 0.2 & 0.2 \\
PPO number of workers & 16 & 32 & 32 & 16 \\
Adam learning rate & 3e-4  & 1e-4 & 1e-4 & 3e-4 \\
Adam $\epsilon$ & 1e-5 & 1e-5 & 1e-5 & 1e-5 \\
PPO max gradient norm & 0.5 & 0.5 & 0.5 & 0.5 \\
PPO value clipping & no & yes & yes & no \\
return normalization & yes  & no & no & yes \\
value loss coefficient & 0.5  & 0.5 & 0.5 & 0.5 \\
student entropy coefficient & 0.0  & 0.0 & 0.0 & 1e-3 \\

\addlinespace[10pt]
\emph{\textbf{PAIRED}} & & \\
generator entropy coefficient & 0.0 & 0.0 & 0.0 & 0.0 \\

\addlinespace[10pt]
\emph{\textbf{HiEnt}} & & \\
student entropy coefficient & 0.05 & 0.005 & 0.005 & - \\
generator entropy coefficient & 0.1 & 0.05 & 0.01 & - \\

\addlinespace[10pt]
\emph{\textbf{BiBC}} & & \\
KL Loss Coefficient & 0.5 & 0.01 & 0.1 & - \\
KL Loss Interval & 10 & 5 & 25 & - \\
KL Loss Direction & bidirectional & bidirectional & bidirectional & - \\
student entropy coefficient & 0.005 & 0.005 & 0.005 & - \\
generator entropy coefficient & 0.05 & 0.005 & 0.05 & - \\

\addlinespace[10pt]
\emph{\textbf{UniBC}} & & \\
KL Loss Coefficient & 0.5 & 0.01 & 0.1 & - \\
KL Loss Interval & 10 & 5 & 25 & - \\
KL Loss Direction & unidirectional & unidirectional & unidirectional & - \\
student entropy coefficient & 0.005 & 0.005 & 0.005 & - \\
generator entropy coefficient & 0.05 & 0.005 & 0.05 & - \\

\addlinespace[10pt]
\emph{\textbf{ACCEL}} & & \\
Edit rate, $q$ & - & 1.0 & - & 1.0  \\
Replay rate, $p$ & - & 0.8 & - & 0.9 \\
Buffer size, $K$ & - & 4000 & - & 1000 \\
Scoring function & - & positive value loss & - & positive value loss \\
Edit method & - & random & - & random \\
Levels edited & - & easy & - & easy \\
Prioritization & -  & rank & - & rank \\
Temperature, $\beta$ & -  & 0.3 & - & 0.1 \\
Staleness coefficient, $\rho$ & - & 0.5 & - & 0.5 \\

\addlinespace[10pt]
\emph{\textbf{PLR}} & & \\
Scoring function & positive value loss & positive value loss & positive value loss & positive value loss\\
Replay rate, $p$ & 0.5 & 0.5 & 0.5 & 0.5 \\
Buffer size, $K$ & 10000 & 4000 & 4000 & 1000 \\

\addlinespace[10pt]
\emph{\textbf{Evo}} & & \\
Scoring function & - & positive value loss  & - &  positive value loss\\
Replay rate (PAIRED+Evo), $p$ & - & 0.5 & - &  0.5 \\
Replay rate (FlexPAIRED+Evo), $p$ & - & 0.9 & - &  0.5 \\
Buffer size, $K$ & - & 4000 & - &  1000 \\

\bottomrule
\end{tabular}%
}
\end{sc}
\end{small}
\end{center}
\vskip -0.1in
\end{table}

\section{Additional results}
\label{sec:additional_results}

Here we report the performance of each design choice on individual test environments. Table \ref{table:cr_mean} shows the mean and standard error on each F1 track (Figure \ref{fig:cr_test_tracks}) when averaged over 10 seeds. Each seed is run for 10 episodes. Similarly, Table \ref{table:mg60_mean} and \ref{table:paired_evo} report the performance on challenging mazes (Figure \ref{fig:mg_test_mazes}) when trained in the Minigrid environment with 0-60 uniform blocks. Note that episodes per level were evaluated in sequence, in contrast to \citet{accel2022}, which evaluated the episodes over several parallel processes. The latter approach is biased toward including shorter episodes over longer episodes. It can be clearly seen that our proposed design choices lead to strong gains in performance over PAIRED, with a high entropy bonus having an especially big impact on the zero-shot generalization capability.


\begin{table}[]
\centering
\resizebox{0.85\columnwidth}{!}{%
    \begin{tabular}{lllllllr}
\toprule
Environment &Robust PLR &PAIRED &HiEnt &UniBC &BiBC &UniBC+Ent &BiBC+HiEnt\\
\midrule
CarRacingF1-Australia&$692.3\pm14.96$&$100.28\pm21.67$&$651.72\pm13.19$&$167.02\pm18.07$&$273.58\pm27.92$&$517.53\pm24.04$&$\mathbf{779.11\pm8.69}$\\
CarRacingF1-Austria&$615.14\pm12.93$&$92.15\pm24.0$&$597.53\pm6.88$&$226.64\pm17.61$&$286.47\pm26.47$&$487.17\pm17.67$&$\mathbf{758.52\pm10.83}$\\
CarRacingF1-Bahrain&$589.83\pm15.05$&$-34.96\pm18.8$&$556.74\pm16.58$&$42.46\pm19.5$&$192.03\pm27.19$&$429.02\pm22.03$&$\mathbf{631.47\pm14.71}$\\
CarRacingF1-Belgium&$473.52\pm12.22$&$72.19\pm20.14$&$452.7\pm6.6$&$143.56\pm16.76$&$229.74\pm22.13$&$301.16\pm16.98$&$\mathbf{587.27\pm15.13}$\\
CarRacingF1-Brazil&$454.92\pm13.35$&$75.73\pm18.16$&$435.26\pm13.2$&$150.35\pm18.02$&$188.18\pm22.24$&$368.09\pm14.52$&$\mathbf{557.48\pm19.77}$\\
CarRacingF1-China&$227.71\pm24.37$&$-100.5\pm9.15$&$180.44\pm30.45$&$-67.78\pm6.95$&$117.33\pm26.1$&$113.43\pm23.97$&$\mathbf{511.67\pm23.78}$\\
CarRacingF1-France&$478.31\pm22.35$&$-80.76\pm12.73$&$482.9\pm28.46$&$61.25\pm20.53$&$187.0\pm29.93$&$299.8\pm22.46$&$\mathbf{603.53\pm15.62}$\\
CarRacingF1-Germany&$498.59\pm17.9$&$-32.51\pm16.05$&$471.62\pm13.22$&$115.9\pm22.96$&$195.58\pm21.53$&$311.54\pm19.91$&$\mathbf{532.02\pm14.86}$\\
CarRacingF1-Hungary&$707.75\pm17.5$&$97.6\pm28.53$&$696.65\pm17.79$&$137.16\pm20.82$&$229.81\pm28.15$&$517.05\pm26.73$&$\mathbf{739.13\pm9.53}$\\
CarRacingF1-Italy&$624.96\pm11.94$&$131.56\pm23.71$&$568.26\pm6.21$&$233.89\pm24.51$&$300.09\pm25.83$&$502.34\pm19.91$&$\mathbf{780.99\pm8.88}$\\
CarRacingF1-Malaysia&$399.86\pm17.51$&$-26.18\pm16.61$&$386.58\pm11.78$&$37.31\pm14.04$&$154.91\pm23.64$&$306.64\pm14.98$&$\mathbf{519.64\pm17.22}$\\
CarRacingF1-Mexico&$\mathbf{712.1\pm12.48}$&$66.53\pm31.4$&$\mathbf{711.34\pm10.05}$&$183.82\pm26.69$&$237.1\pm30.77$&$543.13\pm23.33$&$\mathbf{697.4\pm12.06}$\\
CarRacingF1-Monaco&$485.64\pm19.46$&$-28.3\pm18.18$&$356.1\pm26.13$&$98.85\pm19.73$&$181.65\pm26.54$&$309.81\pm23.53$&$\mathbf{605.03\pm17.37}$\\
CarRacingF1-Netherlands&$419.22\pm25.24$&$70.41\pm20.35$&$464.78\pm17.1$&$135.91\pm18.48$&$231.52\pm23.17$&$312.16\pm23.39$&$\mathbf{612.2\pm17.67}$\\
CarRacingF1-Portugal&$483.44\pm12.7$&$-48.96\pm13.0$&$424.54\pm14.99$&$92.87\pm16.31$&$177.13\pm24.81$&$370.01\pm19.6$&$\mathbf{621.8\pm16.41}$\\
CarRacingF1-Russia&$\mathbf{649.34\pm13.55}$&$51.26\pm20.52$&$603.0\pm22.04$&$202.54\pm22.4$&$248.56\pm26.74$&$452.11\pm24.86$&$\mathbf{668.23\pm12.66}$\\
CarRacingF1-Singapore&$566.37\pm15.11$&$-34.96\pm13.5$&$435.81\pm19.94$&$71.41\pm14.78$&$187.53\pm23.83$&$343.82\pm27.13$&$\mathbf{644.05\pm13.25}$\\
CarRacingF1-Spain&$621.64\pm13.7$&$134.11\pm23.79$&$609.63\pm12.38$&$260.1\pm20.17$&$298.67\pm25.3$&$504.32\pm16.79$&$\mathbf{681.99\pm13.13}$\\
CarRacingF1-UK&$537.58\pm16.81$&$137.63\pm24.67$&$558.9\pm12.65$&$257.21\pm18.42$&$295.4\pm25.75$&$445.71\pm16.96$&$\mathbf{620.75\pm13.1}$\\
CarRacingF1-USA&$380.65\pm33.22$&$-119.17\pm10.99$&$324.71\pm27.69$&$29.12\pm16.07$&$185.31\pm28.58$&$294.41\pm31.42$&$\mathbf{667.87\pm12.87}$\\
\midrule
Mean&$530.94\pm6.67$&$26.16\pm15.22$&$498.46\pm6.3$&$128.98\pm12.13$&$219.88\pm23.08$&$386.46\pm12.29$&$\mathbf{641.01\pm7.21}$\\
        \bottomrule
        \end{tabular}%
}
\caption{Mean and Standard Error averaged over 10 seeds for test returns of each agent on individual F1 track from the \textbf{CarRacing F1 environment}.}
\label{table:cr_mean}
\end{table}

\begin{table}[]
\centering
\resizebox{0.95\columnwidth}{!}{%
    \begin{tabular}{lllllllr}
\toprule
Environment &ACCEL &PAIRED &HiEnt &UniBC &BiBC &UniBC+HiEnt &BiBC+HiEnt\\
\midrule
Labyrinth&$\mathbf{0.73\pm0.18}$&$\mathbf{0.46\pm0.23}$&$\mathbf{0.7\pm0.2}$&$0.36\pm0.18$&$0.02\pm0.02$&$0.33\pm0.1$&$\mathbf{0.48\pm0.2}$\\
Labyrinth2&$\mathbf{0.62\pm0.2}$&$0.15\pm0.14$&$\mathbf{0.67\pm0.18}$&$\mathbf{0.46\pm0.22}$&$0.0\pm0.0$&$0.26\pm0.15$&$\mathbf{0.46\pm0.19}$\\
LargeCorridor&$\mathbf{0.64\pm0.2}$&$0.16\pm0.08$&$\mathbf{0.63\pm0.19}$&$\mathbf{0.62\pm0.16}$&$0.16\pm0.16$&$\mathbf{0.57\pm0.11}$&$\mathbf{0.47\pm0.2}$\\
Maze&$\mathbf{0.79\pm0.2}$&$0.01\pm0.01$&$0.19\pm0.11$&$0.06\pm0.05$&$0.0\pm0.0$&$0.03\pm0.02$&$0.14\pm0.08$\\
Maze2&$\mathbf{0.33\pm0.18}$&$0.02\pm0.02$&$\mathbf{0.54\pm0.16}$&$\mathbf{0.36\pm0.2}$&$0.0\pm0.0$&$\mathbf{0.38\pm0.15}$&$\mathbf{0.24\pm0.19}$\\
Maze3&$\mathbf{0.53\pm0.2}$&$0.15\pm0.15$&$0.39\pm0.19$&$\mathbf{0.8\pm0.2}$&$0.19\pm0.19$&$\mathbf{0.5\pm0.13}$&$\mathbf{0.5\pm0.14}$\\
MiniGrid-FourRooms&$\mathbf{0.51\pm0.04}$&$0.38\pm0.06$&$0.42\pm0.02$&$\mathbf{0.49\pm0.02}$&$0.25\pm0.06$&$\mathbf{0.55\pm0.05}$&$\mathbf{0.51\pm0.04}$\\
MiniGrid-SimpleCrossingS11N5&$\mathbf{0.82\pm0.02}$&$0.48\pm0.16$&$0.62\pm0.04$&$0.65\pm0.03$&$0.27\pm0.14$&$0.68\pm0.08$&$\mathbf{0.86\pm0.04}$\\
PerfectMazeMedium&$\mathbf{0.5\pm0.05}$&$0.17\pm0.07$&$\mathbf{0.47\pm0.15}$&$0.24\pm0.07$&$0.12\pm0.11$&$0.25\pm0.04$&$0.31\pm0.14$\\
SixteenRooms&$0.64\pm0.18$&$0.16\pm0.13$&$0.42\pm0.16$&$\mathbf{0.91\pm0.07}$&$0.2\pm0.18$&$\mathbf{0.83\pm0.07}$&$0.59\pm0.18$\\
SixteenRoomsFewerDoors&$\mathbf{0.55\pm0.19}$&$0.1\pm0.1$&$\mathbf{0.57\pm0.23}$&$0.25\pm0.19$&$0.03\pm0.03$&$\mathbf{0.64\pm0.17}$&$\mathbf{0.41\pm0.17}$\\
SmallCorridor&$\mathbf{0.67\pm0.16}$&$0.25\pm0.17$&$\mathbf{0.63\pm0.22}$&$\mathbf{0.74\pm0.17}$&$0.04\pm0.04$&$\mathbf{0.63\pm0.07}$&$\mathbf{0.4\pm0.19}$\\
\midrule
Mean&$\mathbf{0.61\pm0.03}$&$0.21\pm0.07$&$\mathbf{0.52\pm0.13}$&$0.49\pm0.05$&$0.11\pm0.05$&$0.47\pm0.05$&$0.45\pm0.11$\\
        \bottomrule
        \end{tabular}%
}
\caption{Mean and Standard Error for Solved Rates averaged over 5 seeds of each agent on individual mazes when trained in the \textbf{Minigrid 0-60 Uniform-Blocks environment}.}
\label{table:mg60_mean}
\end{table}

\begin{table}[]
    \centering
\vspace{-3mm}
\resizebox{0.75\columnwidth}{!}{%
    \begin{tabular}{lllr}
    \toprule
         Environment &ACCEL &PAIRED+Evo &FlexPAIRED+Evo \\ 
\midrule
Labyrinth&$\mathbf{0.73\pm0.18}$&$\mathbf{0.51\pm0.28}$&$\mathbf{0.48\pm0.2}$\\
Labyrinth2&$\mathbf{0.62\pm0.2}$&$0.2\pm0.14$&$\mathbf{0.4\pm0.24}$\\
LargeCorridor&$\mathbf{0.64\pm0.2}$&$\mathbf{0.53\pm0.27}$&$0.23\pm0.19$\\
Maze&$\mathbf{0.79\pm0.2}$&$\mathbf{0.5\pm0.27}$&$0.06\pm0.05$\\
Maze2&$\mathbf{0.33\pm0.18}$&$\mathbf{0.28\pm0.24}$&$\mathbf{0.28\pm0.18}$\\
Maze3&$\mathbf{0.53\pm0.2}$&$\mathbf{0.44\pm0.25}$&$\mathbf{0.33\pm0.18}$\\
MiniGrid-FourRooms&$\mathbf{0.51\pm0.04}$&$\mathbf{0.42\pm0.09}$&$0.38\pm0.05$\\
MiniGrid-SimpleCrossingS11N5&$\mathbf{0.82\pm0.02}$&$\mathbf{0.86\pm0.06}$&$0.69\pm0.06$\\
PerfectMazeMedium&$\mathbf{0.5\pm0.05}$&$\mathbf{0.42\pm0.14}$&$\mathbf{0.37\pm0.11}$\\
SixteenRooms&$\mathbf{0.64\pm0.18}$&$\mathbf{0.78\pm0.13}$&$0.26\pm0.16$\\
SixteenRoomsFewerDoors&$\mathbf{0.55\pm0.19}$&$\mathbf{0.48\pm0.23}$&$\mathbf{0.21\pm0.16}$\\
SmallCorridor&$\mathbf{0.67\pm0.16}$&$\mathbf{0.58\pm0.25}$&$\mathbf{0.62\pm0.24}$\\
\midrule
Mean&$\mathbf{0.61\pm0.03}$&$\mathbf{0.5\pm0.14}$&$0.36\pm0.08$\\
    \bottomrule
    \end{tabular}
    }
     \caption{\small{Zero-shot test performance results for PAIRED+Evo and FlexPAIRED+Evo in the test Minigrid mazes when trained in the \textbf{Minigrid 0-60 Uniform-Blocks environment} averaged over 5 seeds and 100 episodes per seed. We report mean and standard error for Solved Rate as the metric here, highlighting in bold the best performing agents.}}
    \label{table:paired_evo}
    \vspace{-3mm}
\end{table}



\begin{figure}[h!]
        \centering
        \includegraphics[width=0.95\linewidth]{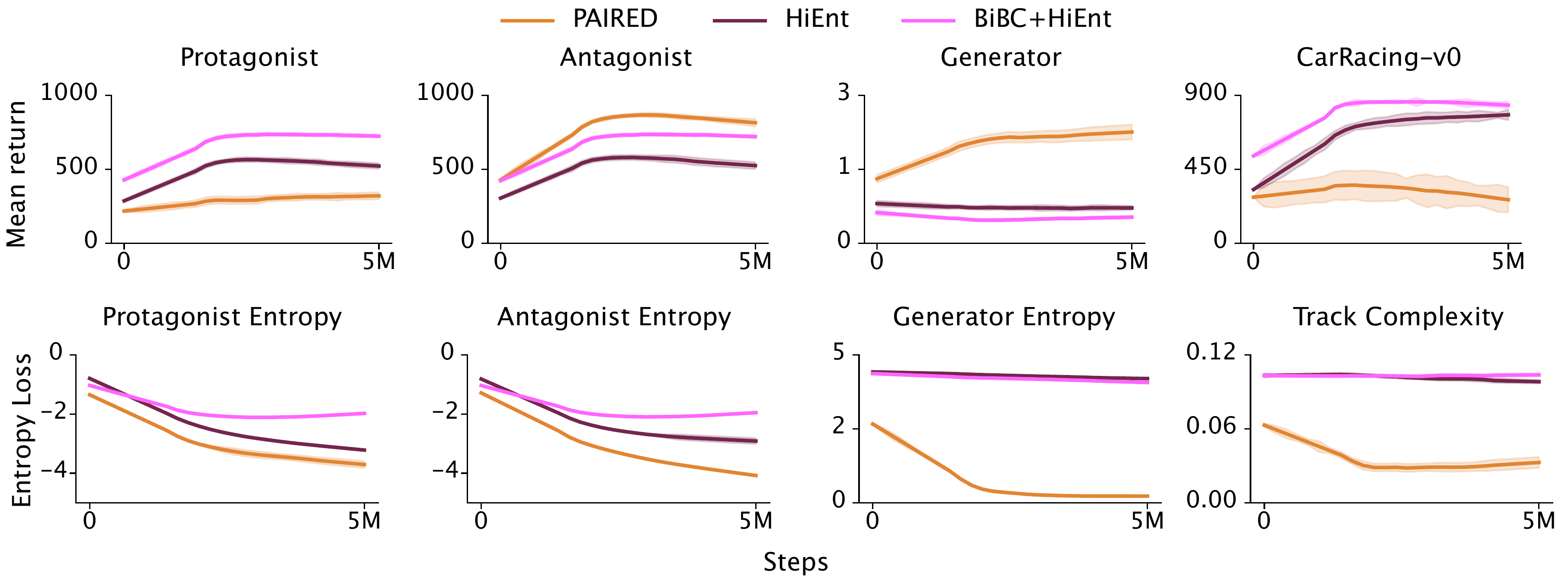}
\caption{Mean agent returns throughout training for all agents in the \textbf{CarRacing F1 environment}.}
\label{fig:cr_train_return}
\end{figure}

Figure \ref{fig:cr_train_return} (top) shows the mean return achieved by the protagonist, antagonist and the adversary throughout training in the generated tracks. We also report the performance of protagonist in the vanilla CarRacing track used for validation purposes. The bottom row shows the entropy loss of each agent as well as the complexity of the tracks generated by the adversary.

\begin{figure}[h!]
        \centering\subfigure[Training returns and Entropy loss]{\includegraphics[width=.5\linewidth]{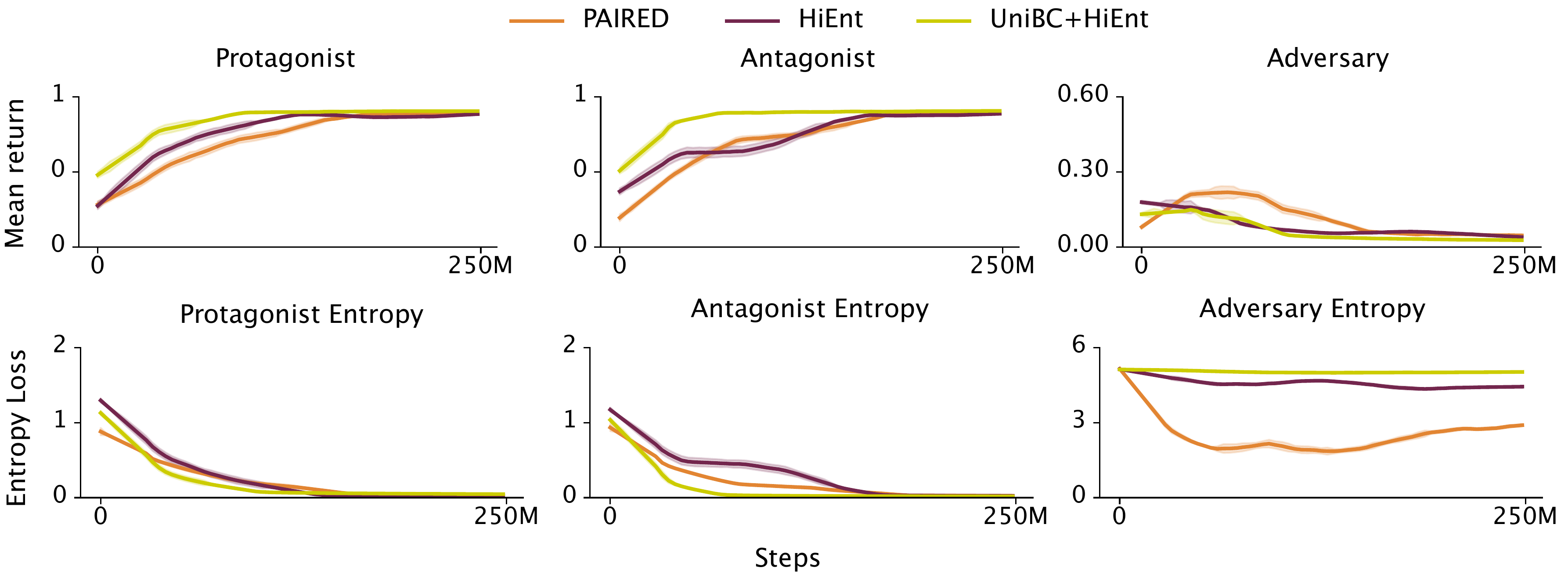}} 
\centering\subfigure[Grid Complexity]{\includegraphics[width=.45\linewidth]{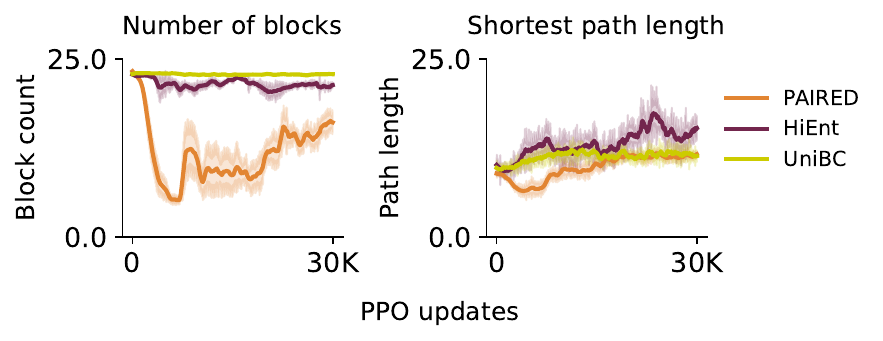}} 
\caption{Mean agent returns throughout training for all agents in the \textbf{Minigrid 25-blocks environment}.}
\label{fig:mg25_train_return}
\end{figure}

\begin{figure}[h!]
\centering
\includegraphics[width=0.85\linewidth]{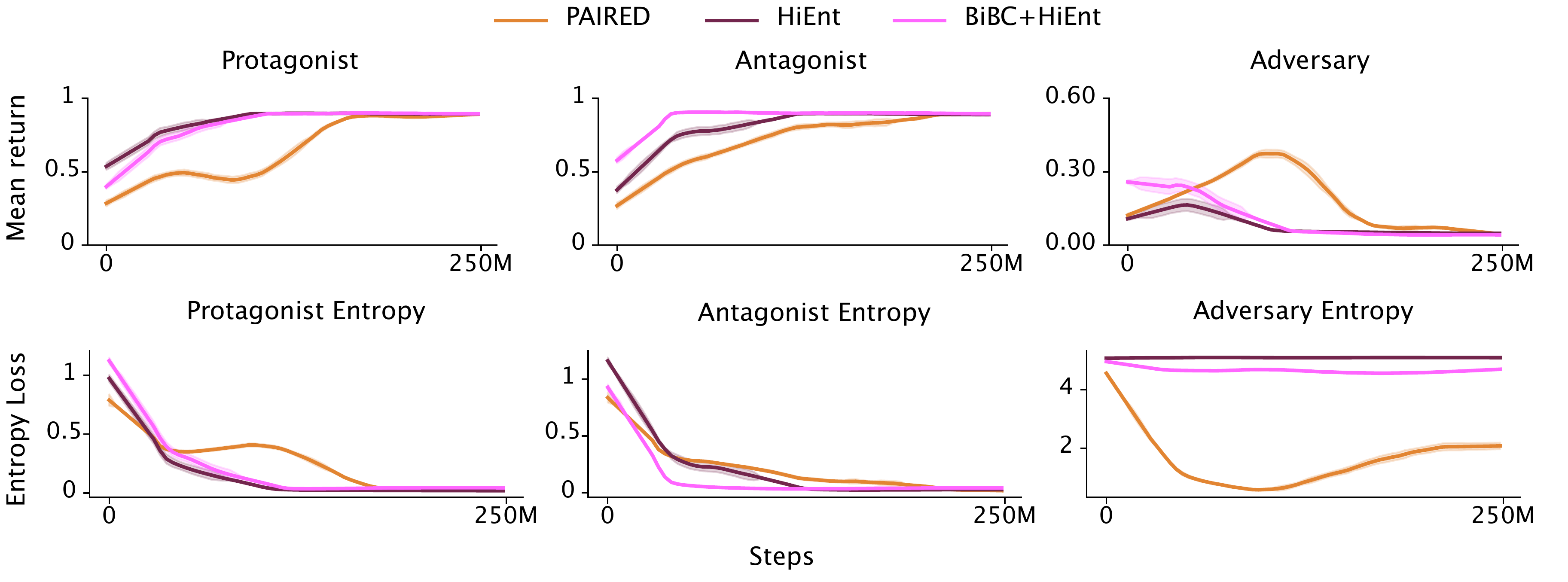}
\caption{Mean agent returns throughout training for all agents in the \textbf{Minigrid [0-60] Uniform-blocks environment}.}
\label{fig:mg60_train_return}
\end{figure}

Similarly, Figure \ref{fig:mg25_train_return} (a) and \ref{fig:mg60_train_return} show the mean returns achieved by the three agents in the generated mazes (top row) and the entropy loss of each agent in those mazes (bottom row). Figure \ref{fig:mg25_train_return} (b) shows the complexity of the mazes generated by the baselines when trained with a fixed budget of 25 blocks. PAIRED experiences a sharp decline in the number of blocks used and hence generates extremely sparse mazes (see Figure \ref{fig:mg25_tracks}), whereas the other two baselines help the adversary in overcoming this problem.

\section{Combining it all!}
\label{app:acp_all}


\begin{figure}[t]
    \centering
    \vspace{-5mm}
    \includegraphics[width=.98\linewidth]{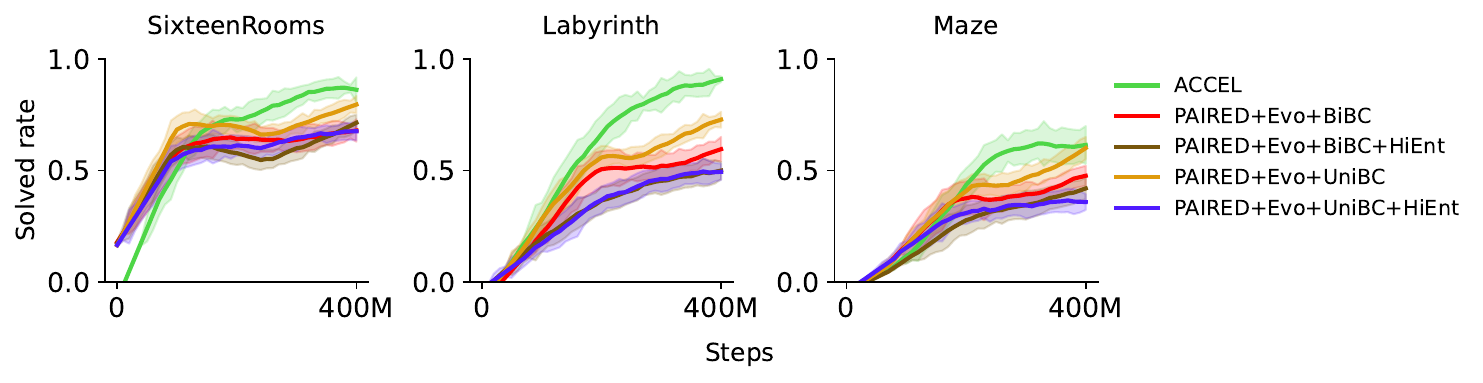}
    \vspace{-5mm}
    \caption{\small{Solved Rates averaged across 10 seeds for PAIRED+Evo+BC+HiEnt and its ablation baselines on Minigrid 0-60 Uniform-blocks environment.}}
    \label{fig:mg60_acp_all_solvedrate}
    \vspace{-5mm}
\end{figure}

In this section, we analyze how all three key design choices - PAIRED+Evo+BC+HiEnt - work together in both the BipedalWalker and Minigrid environments. In Minigrid, we notice that using PAIRED+Evo+UniBC gives the best results for the validation mazes compared to other combinations in Figure~\ref{fig:mg60_acp_all_solvedrate}. When we zero-shot test these design choices on all the 12 mazes, the combination of PAIRED+Evo+UniBC allows PAIRED to match ACCEL in performance on the Maze Benchmark, as shown in Table~\ref{table:mg_acp_all}. We used the default hyperparameters of $KL$ loss coefficient of $0.01$, $KL$ loss interval of $10$, and the generator's entropy bonus to $0.05$ while the student's entropy bonus set to $0.005$.


\begin{table}[h]
    \centering
\resizebox{0.95\columnwidth}{!}{%
    \begin{tabular}{lccccr}
    \toprule
Environment &ACCEL &PAIRED+Evo+BiBC &PAIRED+Evo+BiBC+HiEnt &PAIRED+Evo+UniBC &PAIRED+Evo+UniBC+HiEnt\\
\midrule
Labyrinth&$\mathbf{0.79\pm0.12}$&$\mathbf{0.88\pm0.1}$&$\mathbf{0.75\pm0.1}$&$\mathbf{0.89\pm0.09}$&$0.58\pm0.13$\\
Labyrinth2&$\mathbf{0.82\pm0.09}$&$\mathbf{0.78\pm0.12}$&$\mathbf{0.74\pm0.1}$&$\mathbf{0.92\pm0.08}$&$0.57\pm0.12$\\
LargeCorridor&$\mathbf{0.48\pm0.14}$&$\mathbf{0.5\pm0.13}$&$0.27\pm0.1$&$\mathbf{0.63\pm0.13}$&$\mathbf{0.54\pm0.12}$\\
Maze&$\mathbf{0.81\pm0.11}$&$\mathbf{0.67\pm0.11}$&$0.43\pm0.12$&$\mathbf{0.75\pm0.11}$&$0.48\pm0.14$\\
Maze2&$\mathbf{0.85\pm0.06}$&$\mathbf{0.84\pm0.09}$&$0.62\pm0.09$&$\mathbf{0.88\pm0.07}$&$0.51\pm0.14$\\
Maze3&$\mathbf{0.89\pm0.07}$&$0.62\pm0.12$&$0.43\pm0.1$&$\mathbf{0.89\pm0.07}$&$0.66\pm0.12$\\
MiniGrid-FourRooms&$\mathbf{0.52\pm0.02}$&$\mathbf{0.55\pm0.04}$&$0.48\pm0.03$&$\mathbf{0.55\pm0.03}$&$0.48\pm0.03$\\
MiniGrid-SimpleCrossingS11N5&$0.85\pm0.04$&$0.96\pm0.02$&$0.95\pm0.02$&$\mathbf{0.99\pm0.0}$&$0.95\pm0.02$\\
PerfectMazeMedium&$\mathbf{0.71\pm0.06}$&$0.6\pm0.08$&$0.48\pm0.08$&$\mathbf{0.75\pm0.06}$&$0.49\pm0.1$\\
SixteenRooms&$\mathbf{0.86\pm0.07}$&$0.67\pm0.13$&$\mathbf{0.88\pm0.04}$&$\mathbf{0.91\pm0.05}$&$\mathbf{0.81\pm0.08}$\\
SixteenRoomsFewerDoors&$\mathbf{0.74\pm0.12}$&$0.61\pm0.13$&$\mathbf{0.7\pm0.1}$&$\mathbf{0.86\pm0.08}$&$\mathbf{0.69\pm0.12}$\\
SmallCorridor&$\mathbf{0.72\pm0.13}$&$0.35\pm0.13$&$0.37\pm0.11$&$0.44\pm0.13$&$\mathbf{0.53\pm0.12}$\\
\midrule
Mean&$\mathbf{0.75\pm0.03}$&$0.67\pm0.07$&$0.59\pm0.06$&$\mathbf{0.79\pm0.04}$&$0.61\pm0.08$\\
    \bottomrule
    \end{tabular}
    }
     \caption{\small{Zero-shot test performance results for PAIRED+Evo+BC+HiEnt and its ablations in the test Minigrid mazes when trained in the \textbf{Minigrid 0-60 Uniform-Blocks environment} averaged over 10 seeds and 100 episodes per seed. We report mean and standard error for Solved Rate as the metric here, highlighting in bold the best performing agents.}}
    \label{table:mg_acp_all}
    \vspace{-3mm}
\end{table}

\begin{figure}[H]
    \centering
    \vspace{-3mm}
    \includegraphics[width=.5\linewidth]{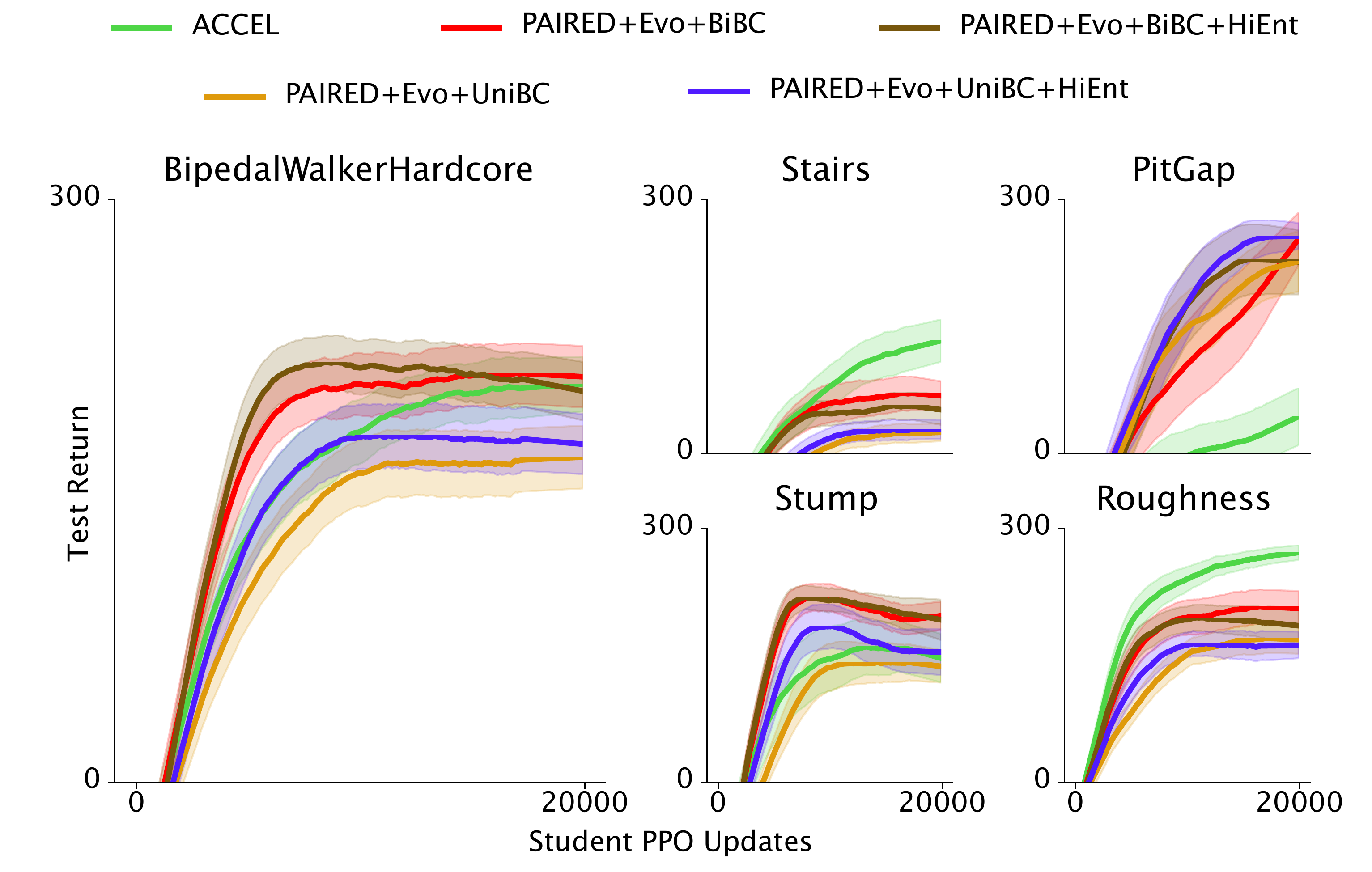}
    \caption{\small{Test returns averaged across 10 seeds for PAIRED+Evo+BC+HiEnt and its ablation baselines on BipedalWalker environment.}}
    \label{fig:bp_acp_all_solvedrate}
    \vspace{-5mm}
\end{figure}

Similarly, in BipedalWalker (Figure~\ref{fig:bp_acp_all_solvedrate}), we observe that ACCEL does better on the Stairs and Roughness challenge, but in PitGap and Stump, PAIRED+Evo+BiBC+HiEnt achieves greater solved rate and better sample efficiency. Also, it is worth noting that in the BipedalWalker-Hardcore environment, the combined method is much more sample efficient as compared to ACCEL. Here we used $KL$ loss coefficient of $0.1$, $KL$ loss interval of $10$, and the generator's entropy bonus set to $0.01$ while the student's entropy bonus set to $0.001$.

These results highlight the effectiveness of our carefully selected design choices and underscore future opportunities for developing better generalizable UED methods.

\section{Psuedocodes}
Here we describe the algorithm of BC framework. Our algorithm is the same as that of PAIRED's \citep{PAIRED}, with an additional KL-divergence term highlighted in blue font. This KL-divergence term can be applied either to both the student agents (BiBC) or to just the protagonist (UniBC), after every $N$ gradient update steps, where $N \in \mathbb{N}$. 
\begin{algorithm}[H]
\caption{PAIRED+BC}
\label{alg:PAIRED_bc}

\begin{algorithmic}[1]

\STATE {\bfseries Input:} Initial policies for protagonist $\pi^P$, antagonist $\pi^A$, initial environment generator $\tilde{\Lambda}$
\WHILE{not converged}
\STATE Use $\tilde{\Lambda}$ to generate environment parameters $\vec{\theta}$
\STATE Collect a trajectory $\tau^P$ using $\pi^P$ in environment $\vec{\theta}$
\STATE Update $\pi^P$ to minimize regret using $\mathcal{L}_{ppo}(\pi_P) + \textcolor{blue}{D_{KL}^{\tau^P}(\pi_A || \pi_P)}$
\STATE Collect a trajectory $\tau^A$ using $\pi^A$ in environment $\vec{\theta}$
\STATE Update $\pi^A$ to maximize regret using $\mathcal{L}_{ppo}(\pi_A) + \textcolor{blue}{D_{KL}^{\tau^A}(\pi_P || \pi_A)}$ \space \space \textcolor{gray}{\# if bidirectional BC}
\STATE Compute the regret as: \\ 
$\normalfont\textsc{Regret}^{\vec{\theta}}(\pi^P, \pi^A) = V^{\vec{\theta}}(\pi^A) - V^{\vec{\theta}}(\pi^P) $ 
\STATE Update $\tilde{\Lambda}$ to maximize regret
\ENDWHILE
\end{algorithmic}
\end{algorithm}

For PAIRED+Evo, we modify ACCEL \citep{accel2022} (highlighted in violet font) to utilize an additional agent which acts as the antagonist, and modify the replay buffer to store or replay levels based on the calculated regret.  

\begin{algorithm}[H]
\caption{PAIRED+Evo}
\label{alg:PAIRED_randomoptim}
\begin{algorithmic}[1]

\STATE {\bfseries Input:} Initial policies for protagonist $\pi^P$, antagonist $\pi^A$, initial environment generator $\tilde{\pi}$, level buffer $\bm{\Lambda}$
\WHILE{not converged}
    \STATE Sample replay decision $d \sim P_{D}(d)$
    \IF{d=0}
        \STATE {\color{gray} \emph{\# Evaluate on levels but do not update}}
        \STATE Use $\tilde{\pi}$ to generate environment parameters $\vec{\theta}$
        \STATE Collect a trajectory $\tau^P$ using $\pi^P$ in environment $\vec{\theta}$ with \textbf{stop gradient}
        \STATE \textcolor{violet}{Collect a trajectory $\tau^A$ using $\pi^A$ in environment $\vec{\theta}$ with \textbf{stop gradient}}
        \STATE \textcolor{violet}{Compute the regret as: \\ 
        $\normalfont\textsc{Regret}^{\vec{\theta}}(\pi^P, \pi^A) = U^{\vec{\theta}}(\pi^A) - U^{\vec{\theta}}(\pi^P) $}
        \STATE Add $\vec{\theta}$ to $\bm{\Lambda}$ if score $S$ meets threshold
    \ELSE
        \STATE {\color{gray} \emph{\# Train on curated high regret levels}}
        \STATE Sample replay level, $\vec{\theta} \sim \bm{\Lambda}$
        \STATE Collect a trajectory $\tau^P$ using $\pi^P$ in environment $\vec{\theta}$
        \STATE Update $\pi^P$
        \STATE \textcolor{violet}{Collect a trajectory $\tau^A$ using $\pi^A$ in environment $\vec{\theta}$}
        \STATE \textcolor{violet}{Update $\pi^A$}
        \STATE {\color{gray} \emph{\# Edit previously high regret levels and evaluate}} 
        \STATE Edit $\vec{\theta}$ to produce $\vec{\theta'}$
        \STATE Collect a trajectory $\tau^P$ using $\pi^P$ in environment $\vec{\theta}$ with \textbf{stop gradient}
        \STATE \textcolor{violet}{Collect a trajectory $\tau^A$ using $\pi^A$ in environment $\vec{\theta}$ with \textbf{stop gradient}}
        \STATE \textcolor{violet}{Compute the regret as: \\ 
        $\normalfont\textsc{Regret}^{\vec{\theta}}(\pi^P, \pi^A) = U^{\vec{\theta}}(\pi^A) - U^{\vec{\theta}}(\pi^P) $}
        \STATE Add $\vec{\theta}$ to $\bm{\Lambda}$ if score $S$ meets threshold
    \ENDIF
\ENDWHILE
\end{algorithmic}
\end{algorithm}

\section{Sample Tracks}
\begin{figure}[h]
\centering\subfigure[PAIRED]{\includegraphics[width=.25\linewidth]{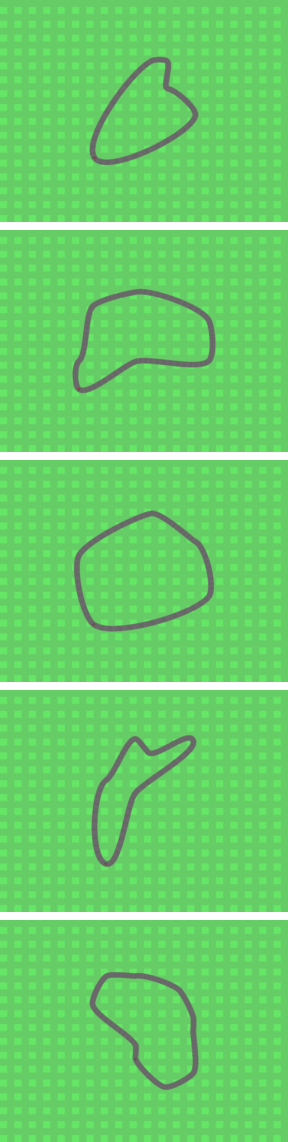}} 
\centering\subfigure[PAIRED+HiEnt]{\includegraphics[width=.25\linewidth]{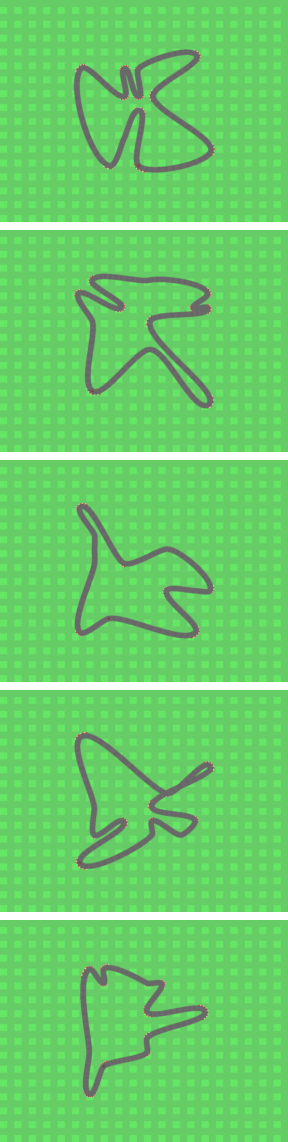}} 
\centering\subfigure[PAIRED+BiBC+HiEnt]{\includegraphics[width=.25\linewidth]{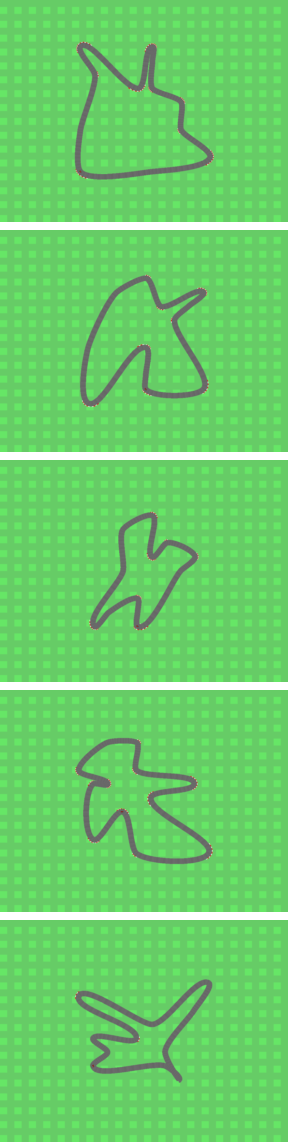}} 
\caption{Sample tracks generated by the adversary in CarRacing Environment.}
\label{fig:cr_tracks}
\end{figure}

\begin{figure}[h]
\centering\subfigure[PAIRED]{\includegraphics[width=.22\linewidth]{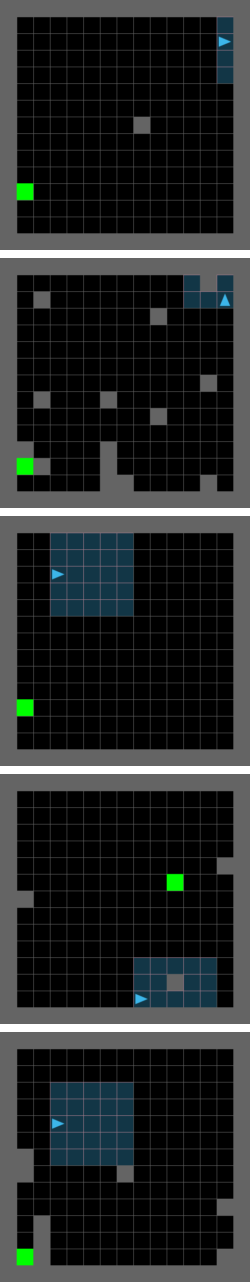}} 
\centering\subfigure[PAIRED+HiEnt]{\includegraphics[width=.22\linewidth]{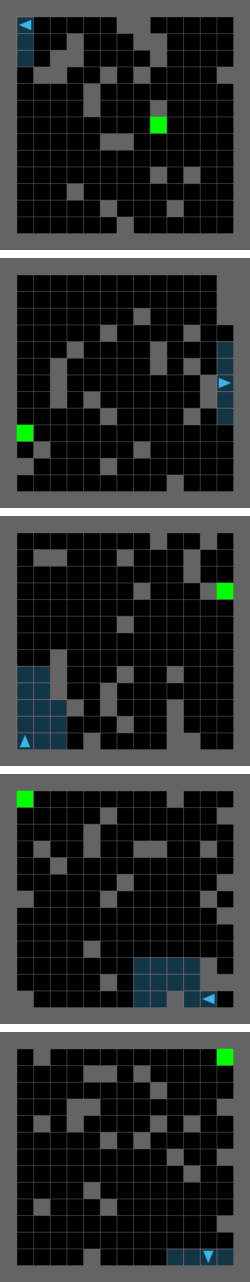}} 
\centering\subfigure[PAIRED+UniBC+HiEnt]{\includegraphics[width=.22\linewidth]{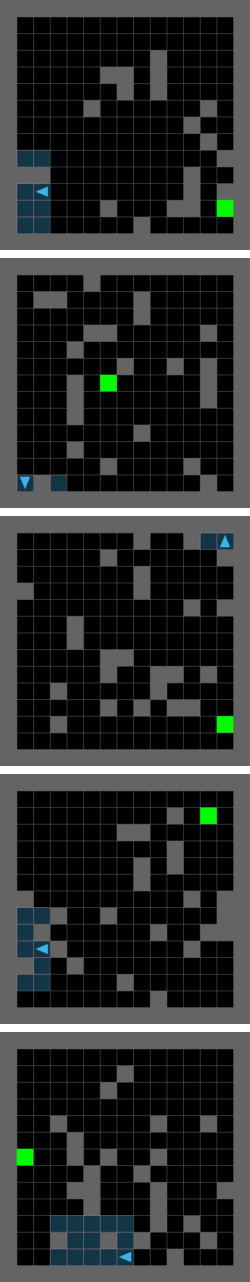}} 
\caption{Sample tracks generated by the adversary in MiniGrid Environment with 25-blocks budget.}
\label{fig:mg25_tracks}
\end{figure}

\begin{figure}[h]
\centering\subfigure[PAIRED]{\includegraphics[width=.22\linewidth]{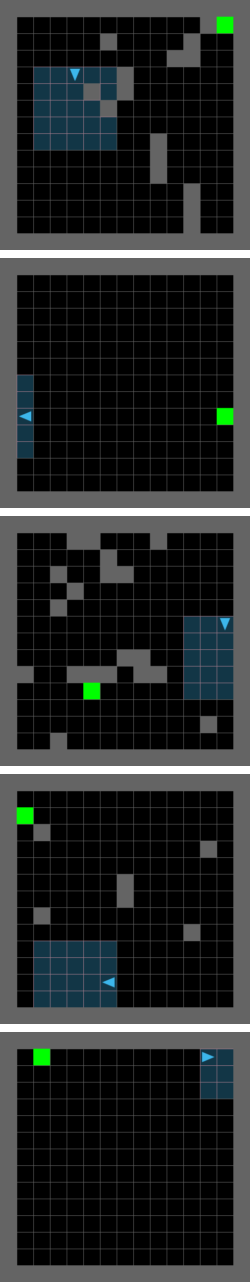}} 
\centering\subfigure[PAIRED+HiEnt]{\includegraphics[width=.22\linewidth]{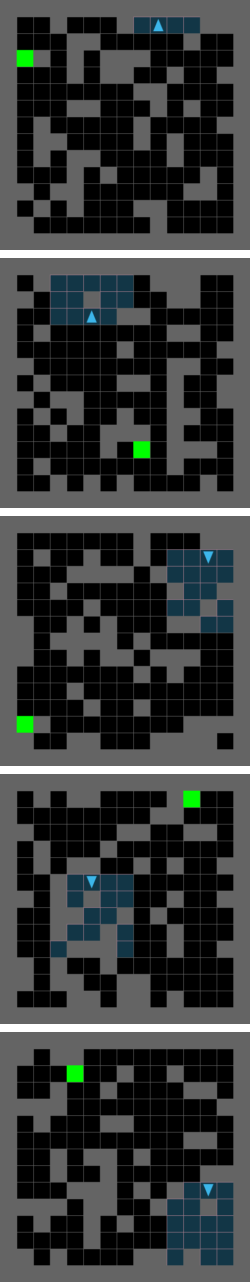}} 
\centering\subfigure[PAIRED+UniBC+HiEnt]{\includegraphics[width=.22\linewidth]{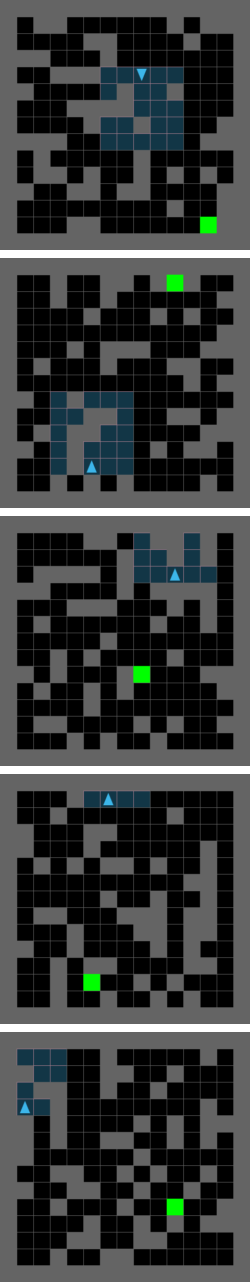}} 
\caption{Sample tracks generated by the adversary in MiniGrid Environment with [0-60] Uniform-blocks budget.}
\label{fig:mg60_tracks}
\end{figure}

Figures \ref{fig:cr_tracks}, \ref{fig:mg25_tracks} and \ref{fig:mg60_tracks} show some sample levels generated by the adversary from each baseline. In CarRacing, PAIRED's adversary keeps on generating simple round tracks which do not transfer to real-world F1 tracks at test time. Similarly, in Minigrid, PAIRED's adversary frequently generates empty mazes which prevent the protagonist from learning navigation skills in these mazes. On the other hand, the adversary in HiEnt and BiBC+HiEnt generates highly complex levels that significantly improve zero-shot generalization. 

\end{document}